\documentclass{article}
\usepackage{mathptmx} %
\usepackage{times} %
\usepackage{amsmath} %
\usepackage{amssymb}  %
\usepackage{booktabs}
\usepackage{graphicx}
\usepackage{adjustbox}
\usepackage[square,numbers,sort&compress]{natbib}
\usepackage{multicol}
\usepackage{multirow}
\usepackage[bookmarks=true]{hyperref}
\usepackage[preprint]{corl_2025} %

\title{ObjectReact: Learning Object-Relative Control for Visual Navigation}

\author{
  Sourav Garg$^{1,\dagger}$ \And 
  Dustin Craggs$^{1,\dagger}$ \And 
  Vineeth Bhat$^2$ \And 
  Lachlan Mares$^1$ \And 
  Stefan Podgorski$^1$ \And 
  Madhava Krishna$^2$ \And 
  Feras Dayoub$^1$ \And 
  Ian Reid$^{1,3}$ \And 
  \vspace{-0.5cm}
  \\
  ${^1}$University of Adelaide, Australia 
  \qquad{}
  ${^2}$IIIT Hyderabad, India 
  \qquad{}
  ${^3}$MBZUAI, UAE 
  \\
  \vspace{-0.25cm}
  \\
  ${^\dagger}$ Equal Contribution
}

\begin{document}
\maketitle

\begin{abstract}
Visual navigation using only a single camera and a topological map has recently become an appealing alternative to methods that require additional sensors and 3D maps. This is typically achieved through an \textit{image-relative} approach to estimating control from a given pair of current observation and subgoal image. However, image-level representations of the world have limitations because images are strictly tied to the agent's pose and embodiment. In contrast, objects, being a property of the map, offer an embodiment- and trajectory-invariant world representation. In this work, we present a new paradigm of learning \textit{object-relative} control that exhibits several desirable characteristics: \textit{a)} new routes can be traversed without strictly requiring to imitate prior experience, \textit{b)} the control prediction problem can be decoupled from solving the image matching problem, and \textit{c)} high invariance can be achieved in cross-embodiment deployment for variations across both training-testing and mapping-execution settings. We propose a topometric map representation in the form of a \textit{relative} 3D scene graph, which is used to obtain more informative object-level global path planning costs. We train a local controller, dubbed \textit{ObjectReact}, conditioned directly on a high-level ``WayObject Costmap'' representation that eliminates the need for an explicit RGB input.
 We demonstrate the advantages of learning object-relative control over its image-relative counterpart across sensor height variations and multiple navigation tasks that challenge the underlying spatial understanding capability, e.g., navigating a map trajectory in the reverse direction. We further show that our sim-only policy is able to generalize well to real-world indoor environments. Code and supplementary material are accessible via project page: \url{https://object-react.github.io/}
\end{abstract}


\section{Introduction}
Navigating in a seen environment is typically accomplished by constructing dense 3D maps, often also using 3D sensors (LiDAR or depth camera). Although capable, these methods still need to rely on rich visual information to effectively understand instructions/goals expressed in natural language. An alternative to 3D based methods is visual topological navigation using only a single camera and a topological map~\cite{savinov2018semi, shah2023vint}, which is often inspired by the navigation abilities of humans. Earlier approaches to topological navigation were mostly limited to teach-and-repeat, which often employed some form of image-based visual servoing to estimate robot velocities given an image pair. Recent methods have proposed to `learn' to predict control signal from the current view and a subgoal image, where the subgoals are generated by a global planner using the topological connectivity of images captured previously from that environment~\cite{savinov2018semi, shah2023gnm}. 

The aforementioned approaches to visual topological navigation can be classified as \textit{image-relative}. Although promising, they remain constrained by their strong reliance on the robot's pose and embodiment when using image subgoals based on an image-level topological world representation. In contrast, objects, being a property of the map, offer an embodiment- and trajectory-invariant world representation. Using an object connectivity-based subgoal representation, we present a new paradigm of learning \textit{object-relative} control  by directly conditioning on the \textit{object subgoals visible in the current image}. This eliminates the concept of having a separate subgoal image and its corresponding limitations of pose and embodiment specificity. As a result, our object-relative controller exhibits more desirable capabilities: \textit{a)} new routes can be traversed, as there is no dependency on retrieving a subgoal image from the prior experience, \textit{b)} control prediction is no longer required to solve an image matching problem between the current RGB and the subgoal image, thus simplifying the learning problem, and \textit{c)} cross-embodiment generalization across the mapping and execution phases can be easily achieved, which would otherwise require quadratically larger datasets to cover different embodiment combinations across the current RGB and the subgoal image observations.

\begin{figure*}
\centering
    \begin{tabular}{cc}
    \includegraphics[width=0.7\textwidth]{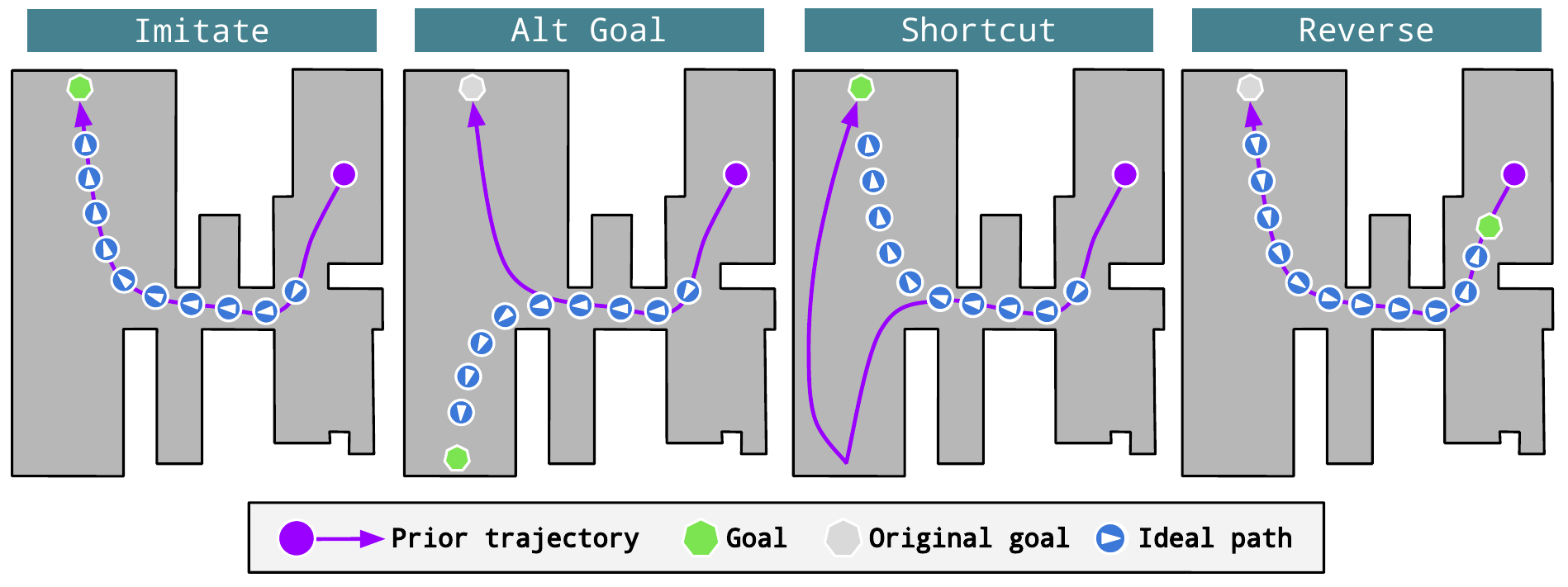}
\\
    \end{tabular}
    \caption{\textbf{Tasks}: Each column shows a topdown view with the prior experience (map) trajectory displayed as a purple path from the purple circle (start) to green point (goal). The tasks are referred to as following: \textit{Imitate} which is akin to teach-and-repeat; \textit{Alt-Goal}, where the goal object is previously seen but unvisited; \textit{Shortcut}, where the prior trajectory is made longer for agent to take a shortcut during inference; and \textit{Reverse}, where agent travels in the opposite direction.
    }
    \label{fig:tasks}
\end{figure*}

We make the following novel contributions towards an object-relative navigation stack: \textit{i)} a \textbf{relative 3D scene graph (3DSG)} representation based on a topometric object-level graph, where inter-image object associations use topological connectivity and intra-image object connectivity is based on relative 3D information; \textit{ii)} a new local controller, dubbed \textbf{ObjectReact}, conditioned on a ``WayObject Costmap'' representation of the object segmentation masks and their path lengths; and \textit{iii)} a \textbf{challenging set of navigation tasks} (see Figure~\ref{fig:tasks} and Section~\ref{sec:exp}) that tests the agent's ability to effectively understand its environment from a limited prior experience.

\def\scaleTaskFig{0.4\textwidth}

\section{Related Work}
Topological approaches to mapping provide an alternative to constructing globally-referenced 3D metric maps for visual navigation. Previous approaches have made extensive use of 3D maps~\cite{ chaplot2020neural, conceptgraphs, salas2013slam++, georgakis2022learning, paul2024lego}, either pre-generated~\cite{huang23vlmaps, weiss2011monocular} or built during exploration~\cite{zhang20233d, zhao2023zero, dorbala2023can, chaplot2020object, TSGM, kwon2021visual}. These map representations also exist in the form of 3DSG~\cite{armeni20193d, rosinol2021kimera}, used for planning~\cite{rana2023sayplan, werby2024hierarchical} and navigation~\cite{ravichandran2022hierarchical, seymour2022graphmapper, singh2023scene, yin2024sg, liu2023bird}. In this work, we focus on a scene graph representation based on intra-frame \textit{relative} 3D object connectivity and inter-frame topological object connectivity, without relying on agent poses, sensor depth or a global frame of reference.

\textbf{Image-Relative Subgoal Control:} In the context of prior map-based navigation, SPTM~\cite{savinov2018semi}, inspired by landmark-based navigation in animals, constructs a topological map with \textit{images} as nodes, and trains a controller to navigate using this map. 
Recent works have expanded on this idea by training a controller that can generalize across multiple embodiments and environments~\cite{shah2023gnm}, perform long-horizon tasks~\cite{shah2022viking}, use goals specified via language~\cite{shah2022lmnav}, and learn to explore when the map is incomplete~\cite{shah2023vint, sridhar2024nomad}. The inter-image edges in these methods are weighted by temporal distance~\cite{shah2021ving, shah2022rapid, savinov2018semi, sutton1988learning}, which is either estimated or deduced. This distance estimator is used to obtain a subgoal image, which is then used in combination with the current observation to obtain a control signal. We refer to this class of methods as \textbf{image-relative}, where a control signal is obtained such that the robot can move from its current position to where the subgoal image from the map was captured. Such controllers (or local trajectory planners) can be based on either action/behavior look-up~\cite{horswill1993polly, matsumoto1996visual, vassallo2000visual}, learning~\cite{thrun1995approach, saxena2017exploring, pathak2018zero, savinov2018semi, li2020learning, meng2020scaling, katara2021deepmpcvs, shah2022lmnav, pathre2024imagine2servo, ehsani2024spoc} or visual servoing~\cite{hutchinson1996tutorial, jones1997appearance, vassallo2000visual, mezouar2002path, blanc2005indoor, remazeilles20063d, cherubini2011visual, diosi2011experimental, bista2016appearance, ahmadi2020visual, feng2021trajectory}.

\textbf{Object-Relative Subgoal Control:} Distinct from image-relative approaches, subgoals can also be defined at a local level based only on the current image observation. Recent works in this direction include RoboHop~\cite{RoboHop}, PixNav~\cite{cai2024bridging}, and TANGO~\cite{podgorski2025tango}, which respectively use objects, a pixel and a 3D point as the subgoal representation in robot's current observation. While these approaches provide a language-based semantic representation that aids in effective global path planning, their local controllers have limitations. RoboHop's zero-shot controller uses fixed linear velocity which is prone to collisions. PixNav's discrete action controller, trained to solve short-horizon navigation using memory-based pixel tracking, tends to overfit to the scene layout. TANGO's occupancy grid-based controller relies on explicit traversability estimation and needs a fallback strategy when traversable regions are out of view. In this work, we focus on obtaining a general, open-set representation of the object subgoals, which is amenable to learning a continuous action controller conditioned only on the currently-viewed objects. We refer to this as an \textbf{object-relative} approach to navigation where its subgoal conditioning variable, being a property of the map, is invariant to the agent's pose and embodiment, unlike its image-relative counterpart.

\textbf{Objects and Semantics for Navigation in Unseen Environments:}
Object-level information in terms of semantic or spatial relations is often used for navigating to a given goal in an unseen environment. BRM~\cite{wu2019bayesian} learns room-level semantic relations to predict subgoals;~\cite{yang2018visual} learns to semantically reason about novel objects; ORG~\cite{du2020learning} specifically learns inter-object spatial correlations;~\cite{ravichandran2022hierarchical} learns a navigation policy on a rich 3DSG representation; TSGM~\cite{TSGM} learns action prediction using a joint image- and object-level representation; and OVG-Nav~\cite{yoo2024commonsense} predicts `object values' using simulator's geodesic distances, as opposed to the use of odometry to predict intra-image distances~\cite{hahn2021no}. However, most of these methods represent graph nodes as images tied to the agent's poses, unlike our object connectivity-based scene graph representation. Moreover, the navigation task for unseen environments inherently focuses on learning to predict a proxy for global path planning, which differs from determining paths through search algorithms~\cite{dijksta1959note} to navigate in seen environments, thus posing different implications on the design choice of underlying representation.

\textbf{Prior Experience-based Navigation:}
Topological navigation has frequently been approached from the perspective of the visual teach-and-repeat task~\cite{furgale2010visual, vsegvic2009mapping, zhang2009robust, dall2021fast, mattamala2022efficient, krajnik2018navigation, halodova2019predictive, do2019high, krajnik2017image, pathak2018zero, kumar2018visual}. These methods use image-based visual servoing to repeat a `teach' run, which is a prior trajectory from the same environment. This approach can be generalized via experiential learning of robot navigation~\cite{levine2023learning}, where a general navigation policy is learned from large, real-world navigation datasets~\cite{shah2023vint}. Such policies can exhibit strong generalization across environments and embodiments, and learn a general understanding of fundamental navigation affordances like traversability, reachability, and exploration~\cite{shah2023gnm, shah2022viking, sridhar2024nomad}. However, such methods are still based on an image-relative control prediction, which has several limitations unlike our proposed object-relative navigation approach. In particular, we decouple our navigation stack into an embodiment-invariant perception module and a prior trajectory-invariant control learning module: our perception system is characterized by its open-set and zero-shot capabilities with the use of models like SAM~\cite{kirillov2023segment, ravi2025sam2, zhao2023fast} and LightGlue~\cite{lindenberger2023lightglue}, whereas our control module is conditioned on a high-level `WayObject Costmap' representation and does not require an explicit RGB input. This unique combination aims to maximize generalization across embodiments and a more challenging set of navigation tasks.

\begin{figure*}
    \centering
    \includegraphics[width=\textwidth]{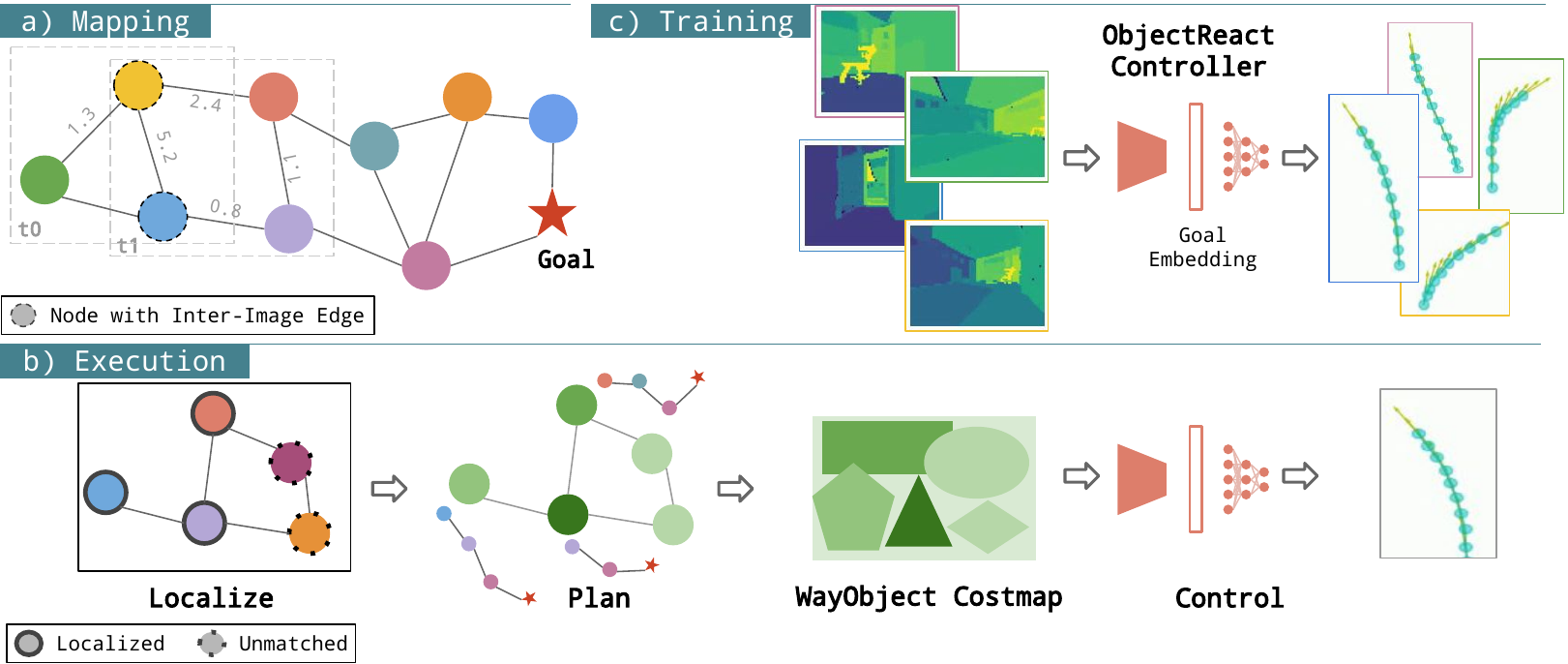}
    \caption{\textbf{Object-Relative Navigation Pipeline}. \textit{a) Mapping:} We construct a topometric map as a \textit{relative} 3D scene graph, where image segments are used as object nodes, which are connected intra-image using 3D Euclidean distances and inter-image using object association.\textit{ b) Execution:} Given the map, we localize each of the query objects and compute its path to the goal node; we assign these path lengths to the object's segmentation mask, forming a ``WayObject Costmap'' for control prediction. \textit{c) Training:} We train a model to learn an ``ObjectReact'' controller that predicts trajectory rollouts from WayObject Costmaps.}
    \label{fig:pipeline}
\end{figure*}

\section{Approach}
Our object-relative navigation pipeline revolves around a representation of the world in the form of object-level connectivity. There exist different ways in which this connectivity can be established. It can be characterized in terms of the topology being flat~\cite{RoboHop} versus hierarchical~\cite{armeni20193d}, and connections that range from being purely topological~\cite{RoboHop} to being fully metric~\cite{conceptgraphs}.
In this work, we build on RoboHop's~\cite{RoboHop} object-level topological mapping. RoboHop connects objects within an image through a 2D Delaunay triangulation of the objects' pixel centers. Such connectivity is prone to geometric ambiguity which affects the shortest-path computation during planning, thus potentially misguiding the controller. We address this limitation through a more informative \textit{relative 3D connectivity}. Using this novel map representation, we use RoboHop's object-level localization and planning to obtain object-level paths to the goal. While RoboHop uses a zero-shot proportional yaw controller, in this work, we propose a controller that \textit{learns to react to objects}. This uses the object-level path lengths in the form of a dense representation, dubbed \textit{WayObject Costmap}, to predict a trajectory rollout. In the following sections, we describe our object-relative navigation pipeline in terms of three distinct phases: mapping, execution, and training, as also explained in Figure~\ref{fig:pipeline}.

\subsection{Mapping Phase: \textit{Relative} 3D Scene Graph}
\textbf{Image Segments as Object Nodes}:
Given a prior map in the form of an image sequence, we use a foundational model, such as SAM2~\cite{ravi2025sam2} or FastSAM~\cite{zhao2023fast}, to extract open-set, semantically-meaningful segmentation masks. For each image segment, we represent an object node in a graph $\mathcal{G}$ using its 2D binary segmentation mask array $\mathbf{M}_i$ and a 3D coordinate $\mathbf{p}_i$, which is defined as the farthest point on the object in a local frame of reference. Unlike 3D mapping techniques that simultaneously estimate camera pose and global scene geometry, we are only interested in a \textit{relative 3D layout of objects} in an image. We use monocular depth estimation~\cite{depthanything} and an arbitrary but fixed focal length to project the 2D object pixels into 3D.

\textbf{Intra-image Edges}: Given the object nodes in an image, we create intra-image edges using \textit{all} pairs of objects. We use the Euclidean distance $e_{ij}$ between the 3D coordinates of the object pairs as edge weights. This relative 3D intra-image connectivity is much more informative in global path planning than a purely 2D connectivity based on object pixel centers, e.g., that used in ~\cite{RoboHop}. We ablate these two types of connectivity in Section~\ref{sec:results_ablations}.

\textbf{Inter-image edges}: We create inter-image edges by tracking objects across consecutive frames using local feature matching. Specifically, we extract SuperPoint~\cite{detone2018superpoint} keypoints and descriptors, and match them using LightGlue~\cite{lindenberger2023lightglue} to identify pixel-level correspondences between an image pair. Using this, segment-level correspondences are established on the basis of the number of pixel-level correspondences that fall into that segment pair. For these inter-image edges, we use an edge weight of $0$ (equivalent to merging nodes) so that the global planner does not incur any cost on traversing these edges to find the shortest path to the goal.

\subsection{Execution Phase: Object Localizer, Global Planner, and Local Controller}
During task execution, we follow RoboHop's object localization and global path planning method. Using the aforementioned segment matching approach, we obtain map object nodes that match with the query object nodes. The global planner computes path lengths from these matched map nodes to the \textit{long-horizon} goal node using Dijkstra's algorithm~\cite{dijksta1959note}. For a query node having multiple matches, the map node with the least path length is selected. Using these path lengths and the segmentation masks, we define a novel representation to learn a local controller which predicts a trajectory rollout, as described in the following section.

\subsection{Training Phase: The \textit{ObjectReact} Controller}
\label{method:train}
One of the most common ways of supervising a goal-conditioned control policy is to predict a control signal given a pair of current RGB image and a subgoal image~\cite{savinov2018semi,shah2023gnm}, similar to image-based visual servoing~\cite{remazeilles20063d, hutchinson1996tutorial}. In this work, we represent subgoals in the form of a \textit{WayObject costmap}, obtained from the path lengths and the segmentation masks of each of the visible objects in the current RGB image. This forms an interpretable representation in terms of a visual distribution of attractiveness of the objects so that a model can learn to react to these attracting (where to go) and repelling (where not to go) objects. 

\textbf{WayObject Costmap}: Since our aim is to learn a local controller conditioned on object-level costs, we consider the following aspects to represent our conditioning variable: i) \textit{Local} Cost Distribution: the path lengths obtained from the planner are the absolute global costs; this can bias training in terms of episode lengths or distance to final goal, and it may not generalize across different types of edge weights used for planning (e.g., 2D vs. 3D). We address this by normalizing the costs per image, as intuitively only a local distribution of \textit{subgoal} costs should be needed to predict a control signal. ii) Object's \textit{Spatial} Distribution: we need to represent the objects themselves while considering a variable number of objects for every image. To address this, we combine the segmentation masks of the objects and their path lengths to form a multi-channel image, dubbed \textit{WayObject Costmap} $\mathcal{W}_{H \times W \times D}$, where a pixel value corresponds to the path length of the object that it belongs to. Instead of directly using the scalar costs, we use a $D$-dimensional encoding $E$ to represent the cost. This is obtained by rescaling the costs and converting them into sine-cosine embeddings, similar to the positional encoding used in transformer architectures~\cite{vaswani2017attention}. Please refer to Appendix Section~\ref{sec_app:costmap} for further details. Thus, $\mathcal{W}$ not only accommodates a variable number of objects per image but also retains spatial and semantic information. This also helps to eliminate the need for an explicit RGB input for learning control, as demonstrated in the ablation studies in Section~\ref{sec:results_ablations}.

\textbf{Trajectory Prediction}:
In this work, we aim to establish distinctive properties of object-relative navigation in contrast to image-relative navigation. Thus, we adapt an existing image-relative controller's training pipeline, which enables direct ablative analyses to clearly highlight the benefits of our proposed approach. We use GNM~\cite{shah2023gnm} for this purpose with two key variations: a) we use HM3D's photorealistic environment to generate training data, and b) we use a custom goal encoder to facilitate multi-channel input of the WayObject Costmap $\mathcal{W}_{H \times W \times D}$. We provide details for the model architecture and loss functions in the Appendix Section~\ref{sec_app:model}.

\textbf{Training Data}:
We use Habitat-Matterport 3D dataset (HM3Dv0.2)~\cite{Ramakrishnan2021HabitatMatterport3D} to train and evaluate our proposed method and the image-relative baseline GNM~\cite{shah2023gnm}. Specifically, we use the training and validation set of the InstanceImageNav challenge set (IIN-HM3D-v3)~\cite{krantz2022instance}. For each of the $145$ unique scenes in the IIN-train set, we uniformly sample $20$ episodes. For each episode, we use its start and end agent states to compute the shortest path using the simulator's geodesic distance estimator. We convert these 2D path coordinates into an interpolated 3D trajectory composed of pure translations of $0.2m$ and pure rotations of $15^\circ$. For training the controller, we further split the IIN-train set to create $80/20$ train/val sets. For generating WayObject Costmaps during training, we use the object instances and depth from the simulator to create a topometric map. As we normalize the path lengths per image, our controller trained with simulator-based costmaps is still able to generalize to the costmaps obtained from inferred segmentation, matching, and depth during mapping. There also exist image segments that were either not detected, mismatched during localization (during execution phase), or did not have a valid path length; we mark these as outliers in the costmap by using a fixed high cost. To further improve generalization from training costmaps to deployment costmaps, we perform a data augmentation where the cost of $30\%$ of the segments is replaced by the outlier cost at random.

\section{Experimental Setup}
\label{sec:exp}
\textbf{Evaluation Dataset}: We use the IIN-val set to evaluate navigation performance. It has $36$ unique scenes from which we sample one episode each. For each episode, we compute a 3D trajectory as described earlier, which serves as the prior map. In the offline mapping phase, we compute a relative 3DSG using RGB images and relative monocular depth. For evaluating navigation performance, i.e., the online execution phase, we initialize the agent along the map trajectory such that it is at least (geodesically) $5m$~\cite{cai2024bridging} away from the goal -- this involves crossing multiple rooms and corridors on the same floor of the house. In all our experiments, the agent is given access to ground truth localization in the form of map image index which is closest to the agent's current 2D position.

\textbf{Evaluation Metrics}:
We evaluate a controller's ability to navigate to an object goal in a given episode. We use an oracle stop condition, i.e., an episode is deemed successful if the robot reaches within $1m$~\cite{habitatchallenge2023} of the goal position in maximum $300$ steps. We report Success weighted by Path Length (SPL)~\cite{anderson2018evaluation} and Soft-SPL (SSPL)~\cite{datta2021integrating}. The latter is particularly useful for episodes that are deemed failures (SPL$=0$) but have progressed toward the goal (SSPL$>0$). We report these metrics as an average over $72$ runs, as each of the $36$ episodes is tested for two different sensor heights in the execution phase: $1.3m$, representing a mobile manipulator such as Stretch~\cite{kemp2022designstretchcompactlightweight}, and $0.4m$, representing a quadruped robot such as Go1~\cite{unitree_go1}. We use a fixed sensor height of $1.3m$ for map images to test the robustness to cross-embodiment deployment across mapping and execution.

\textbf{Tasks}: As illustrated in Figure~\ref{fig:tasks}, we consider four unique navigation tasks: \textit{Imitate}, where agent imitates its prior trajectory akin to teach-and-repeat; \textit{Alt Goal}~\cite{podgorski2025tango}, where an agent needs to visit a previously seen but unvisited goal, thus having to traverse a new route; \textit{Shortcut}, where the prior mapping trajectory has an additional stop at the Alt Goal, so the agent must take a shortcut to the final goal; and \textit{Reverse}, where the agent is tested for its ability to travel in the opposite direction of its prior trajectory. We exclude some episodes for certain tasks if the object goal or the path is found to be invalid upon manual inspection, as described in the Appendix Section~\ref{sec_app:excluded_episodes}.

\section{Results and Discussion}

\subsection{Image-Relative vs. Object-Relative}
Table~\ref{tab:res_main} ablates image-relative and object-relative approaches to learning \textit{subgoal-conditioned} control. We report SPL and SSPL on four navigation tasks, averaged over two different sensor heights. The image-relative baseline is the GNM~\cite{shah2023gnm} model trained on the same HM3D data as our object-relative model ( Section~\ref{sec_app:quant} in the appendix provides a comparison against off-the-shelf GNM model). On the \textit{Imitate} task, both the image-relative and the object-relative methods perform similarly. However, for the other more challenging tasks, the object-relative controller achieves much higher performance than its image-relative counterpart. For the \textit{Alt Goal} task, the image-relative approach suffers mainly from its reliance on the prior experience of having captured an image of the object goal from close proximity. In contrast, the trajectory-invariant nature of our object-level representation enables it to directly reach previously unvisited objects. The \textit{Shortcut} task highlights a different weakness of the image-relative navigation pipelines: their subgoal \textit{image} selection is based on `temporal' distance prediction~\cite{sutton1988learning, savinov2018semi, shah2021ving} which may not accurately capture geometric or geodesic distances, thus missing potential shortcuts. On the other hand, the object-to-goal distances in our object-relative navigation pipeline are already grounded in geometry; thus, it mainly needs to perform accurate object association, which benefits from our modular design that enables drop-in replacement of robust perception techniques. On the \textit{Reverse} task, yet another type of weakness emerges for image-relative methods in terms of their ability to recognize places from an opposite viewpoint~\cite{garg2018lost, keetha2023anyloc}. This challenging data association task adversely affects both the temporal distance prediction and the waypoint prediction. Although the same level of challenge applies to the object-relative controller, it benefits from its `partial' matching~\cite{garg24revisitanything}; that is, even if a small number of objects is matched, it could be sufficient to point the agent in the right direction.

Overall, these results show that an object-relative navigation pipeline addresses multiple weaknesses of its image-relative counterpart, which makes it inherently capable of solving different types of navigation tasks. However, we note that absolute performance on the more challenging tasks is much inferior to the simple \textit{Imitate} (teach-and-repeat) task. We observed that it is largely attributed to lack of robust perception rather than the learnt controller. This is demonstrated in Section~\ref{sec:results_ablations}, which shows that high performance can be achieved across the board when using the object instance ground truth from the simulator for segmentation and matching.

\begin{table*}
\centering
\caption{Comparing image-relative and object-relative controllers across four navigation tasks. 
}
\begin{adjustbox}{width=0.9\textwidth}
\begin{tabular}{l cc cc cc cc} 
\toprule
 {\textbf{Method}} & \multicolumn{2}{c}{\textbf{Imitate}} & \multicolumn{2}{c}{\textbf{Alt Goal}} & \multicolumn{2}{c}{\textbf{Shortcut}} & \multicolumn{2}{c}{\textbf{Reverse}}  \\
 \midrule
   & SPL & SSPL & SPL & SSPL &SPL & SSPL & SPL & SSPL \\
\cmidrule(lr{0.75em}){2-3}
\cmidrule(lr{0.75em}){4-5}
\cmidrule(lr{0.75em}){6-7}
\cmidrule(lr{0.75em}){8-9}

Image Relative: GNM~\cite{shah2023gnm}  & 57.58 & \textbf{66.15} & 2.17 &  13.54 & 7.69 &  18.40 & 11.60 & 23.59 \\

Object Relative: ObjectReact (Ours) & \textbf{59.08} & 64.62 & \textbf{21.74} & \textbf{27.40} & \textbf{23.08} & \textbf{39.56} & \textbf{26.67} & \textbf{36.69}\\

 \bottomrule
\end{tabular}
\end{adjustbox}
    \label{tab:res_main}
\end{table*}

\begin{table}
    \centering
    \caption{Effect of \textbf{Embodiment (height)} variations during execution for fixed map height of 1.3m.}
\begin{adjustbox}{width=0.65\textwidth}
    \begin{tabular}{l cc cc cc}
    \toprule
        \textbf{Method / Robot Height} & \multicolumn{2}{c}{\textbf{0.4m}} & \multicolumn{2}{c}{\textbf{1.3m}} & \multicolumn{2}{c}{\textbf{$|\Delta| \,$} $(\downarrow$ better$)$} \\

        \midrule
        
         & SPL & SSPL & SPL & SSPL & SPL & SSPL \\
        \cmidrule(lr{0.75em}){2-3}
        \cmidrule(lr{0.75em}){4-5}
        \cmidrule(lr{0.75em}){6-7}

Image Relative & 33.33 & 45.93 & 81.82 & 86.38 & 48.49 & 40.45 \\
Object Relative & 60.60 & 68.51 & 57.56 & 60.72 & \textbf{3.04} & \textbf{7.79} \\

    \bottomrule
    \end{tabular}
    \end{adjustbox}
    \label{tab:embodiment}
\end{table}

\subsection{Embodiment Variations based on Robot Height}
Embodiment variations can occur in several ways, making it harder to learn a controller that can generalize well. Even a common variation, such as the height of a robot, can easily trigger failures. One common solution for this is to train on a diverse set of embodiments to learn invariance, which is typically achieved by collecting real-world trajectories from many different robots~\cite{shah2023gnm}. Although this enables cross-embodiment deployment across train-test variations, it does not necessarily account for variations across mapping and execution phases. The latter is particularly important for the \textit{reuse of prior map experience} across different robots. It is worth noting that learning mapping-execution invariance requires \textit{trajectory association across different robots in the exact same environment} -- this not only increases the training data requirement quadratically, but also increases manual effort for annotation. Our proposed object-relative navigation pipeline addresses this problem through its object-level world representation. This leads to high invariance to both the robot's embodiment and prior trajectory while isolating the image matching problem from the control learning problem, thus eliminating the need for associating trajectories across embodiments. 

In Table~\ref{tab:embodiment}, we study the effect of sensor height variations across mapping and execution runs. It can be observed that the image-relative approach performs really well when the map height ($1.3m$) matches the execution height. However, when the execution height is changed to $0.4m$, this performance drops by an absolute $48\%$ (SPL). On the other hand, our object-relative approach is almost invariant to such variations, where it seems to slightly benefit from a lower height likely due to an increased visibility of immediately traversable regions.

\begin{table}
\centering
\caption{\textbf{Ablation Studies} for different types of controllers, map/edges, and conditioning inputs.}
\begin{adjustbox}{width=0.8\textwidth}

\begin{tabular}{lc cc cc cc cc} 
\toprule
  \textbf{Method} &\multicolumn{2}{c}{\textbf{Imitate}} & \multicolumn{2}{c}{\textbf{Alt Goal}} & \multicolumn{2}{c}{\textbf{Shortcut}} & \multicolumn{2}{c}{\textbf{Reverse}} \\

\midrule

   & SPL & SSPL & SPL & SSPL &SPL & SSPL & SPL & SSPL \\

\cmidrule(lr{0.75em}){2-3}
\cmidrule(lr{0.75em}){4-5}
\cmidrule(lr{0.75em}){6-7}
\cmidrule(lr{0.75em}){8-9}

   PixNav~\cite{cai2024bridging} & 42.42 & 51.78 & 45.65 & 54.94 & 19.23 & 33.04 & 26.41 & 39.14 \\

 RoboHop~\cite{RoboHop}  & 63.52 & 73.18 & 45.64 & 65.38 & 48.03 & 57.88 & 48.28 & 66.97  \\

   (Ours) ObjectReact w. 2D Edges & 59.08 & 71.61 & 52.17 & 64.83 & 19.23 & 32.12 & 50.00 & 64.58 \\
  
  (Ours) ObjectReact w. RGB & 63.61 & 75.08 & \textbf{54.35} & \textbf{71.97} & 50.00 & 63.24 & 56.54 & 68.17 \\
  
  (Ours) ObjectReact & \textbf{71.20} & \textbf{82.96} & \textit{54.32} & \textit{70.89} & \textbf{59.61} &\textbf{ 70.40} &\textbf{ 66.66} & \textbf{73.77}\\

 \bottomrule
\end{tabular}
\end{adjustbox}
    \label{tab:ablations_all}
\end{table}

\subsection{Ablation Studies}
\label{sec:results_ablations}
\textbf{Controller Types:}
In Table~\ref{tab:ablations_all}, we compare different object-level controller methods: a) \textit{RoboHop}~\cite{RoboHop} -- a zero-shot controller with a \textit{fixed} linear velocity and a proportional yaw control based on a cost-weighted average of the 2D object centers, and b) \textit{PixNav}~\cite{cai2024bridging} -- a discrete action controller trained in HM3D to track pixel goals (see Appendix Section~\ref{sec_app:baselines} for baseline implementation details). To observe the controller performance in isolation, we use the WayObject Costmaps generated using the object instance ground truth and relative 3D edges-based map connectivity. Thus, both RoboHop and PixNav are provided the same costmap input as ObjectReact for a fair comparison. It can be observed that our proposed method outperforms both PixNav and RoboHop across the board by large margins.

\textbf{Topological (2D) versus Topometric (3D) Maps:} 
In Table~\ref{tab:ablations_all}, we ablate two types of \textit{intra-image} edges for object connectivity in the map. ObjectReact uses 3D distances between all object pairs to define intra-image edges. We compare it with 2D Delaunay Triangulation-based intra-image edges (middle row in Table~\ref{tab:ablations_all}), as proposed in RoboHop~\cite{RoboHop}. It can be observed that 3D information, even in the form of a \textit{relative} 3DSG, is much more informative than its 2D counterpart.

\textbf{Subgoal conditioning with and without current RGB:}
Unlike image-relative controllers that require both the current RGB image and the reference subgoal image, our object-relative controller only relies on the object subgoals (WayObjects) present \textit{within} the current RGB observation. This uniquely allows us to learn control solely from the WayObject Costmap, without requiring an explicit RGB input. We ablate this aspect of our controller in Table~\ref{tab:ablations_all}, which shows that better generalization is achieved \textit{without} the RGB input. 
We speculate that since the current RGB image alone does not contain explicit information about the goal, an over-reliance on it leads to overfitting on specific object instances in the training dataset. WayObject Costmap, on the other hand, provides a representation that is largely invariant to the exact visual appearance of scenes and objects.

\section{Conclusion and Future Work} 
\label{sec:conclusion}
In this work, we proposed an object-relative navigation pipeline with a novel map representation based on a relative 3D Scene Graph, and a new learnt local controller conditioned only on the currently-viewed objects represented in the form of a WayObject Costmap. We demonstrated that the type of subgoal conditioning to learn a local controller has strong implications on its ability to solve challenging tasks beyond teach-and-repeat and cross-embodiment generalization, particularly that across mapping and execution phases. We showed that it is possible to achieve such capabilities by learning an object-relative controller that inherently addresses the limitations of an equivalent image-relative controller. We presented several ablation studies to show the efficacy of our approach while also reporting key limitations, e.g., lack of robust perception techniques that are part of our modular navigation stack. Future work can further extend the capabilities of our method through its WayObject Costmap representation, which is a generic conditioning variable that can be generated through alternative sources, e.g., language instructions, exploration objectives, cross-sensor modalities, or directly from a vision LLM. Furthermore, our object-relative navigation pipeline is closer to the landmark-based navigation strategies observed in human trials~\cite{foo2005humans}; future research in this direction could bridge the gap between the visual navigation capabilities of humans and robots.   

\acknowledgments
We thank Nuri Kim for her valuable insights and discussions. Thanks to Mehdi Hosseinzadeh, Xiangyu Shi, and Hui Wang for their support with initial prototyping. Thanks to Krishan Rana for encouraging discussions. This work was supported with supercomputing resources provided by the Phoenix HPC service at the University of Adelaide.

\appendix
\section*{Appendix}
Here, we provide additional details related to method implementation, experimental setup, and limitations. We also include additional results and analyses, especially in reference to real-world and simulator demonstration videos on our project page: \url{https://object-react.github.io/}

\section{Implementation Details}
\subsection{WayObject Costmap Representation}
\label{sec_app:costmap}
We elaborate on our proposed learnable representation of the segmentation masks and their corresponding topometric path lengths. Given a \textit{variable} number ($N_m$) of binary object masks as a tensor $\mathbf{M} \in \{0,1\}^{H \times W \times N_m}$ in the current image and their scalar path lengths $l$ to the long-horizon goal, we obtain the path length encoding $\mathbf{E} \in \mathbb{R}^{N_m \times D}$ and the WayObject Costmap representation $\mathcal{W} \in \mathbb{R}^{H \times W \times D}$ as below:

\begin{equation}
    \mathcal{W}_{H \times W \times D} = \mathbf{M}_{H \times W \times N_m} @ \hspace{0.1cm} \mathbf{E}_{N_m \times D}
\end{equation}
\begin{equation}
E(l)_i = 
\begin{cases}
\sin\left( \dfrac{l}{Z^{i/D}} \right) & \text{if } i \text{ is even} \\
\cos\left( \dfrac{l}{Z^{(i-1)/D}} \right) & \text{if } i \text{ is odd}
\end{cases} \quad \forall i \in [0,D]
\label{eq:encoding}
\end{equation}
where $\mathbf{M}$ represents $N_m$ segmentation mask arrays of height $H$ and width $W$, corresponding to the currently-viewed objects. $\mathbf{E}$ represents a $D$-dimensional positional encoding~\cite{vaswani2017attention} of the path lengths for each of the objects. The path lengths are normalized per image and re-scaled before encoding such that $l \in [1,L]_\mathbb{Z}$. $l=L$ represents the shortest path length per image and $l=0$ represents outliers corresponding to undetected and unmatched image segments. We set $D=8$, $Z=10000$, and $L=100$.

\subsection{Model Architecture and Loss functions} 
\label{sec_app:model}
We base the training of our navigation controllers on GNM~\cite{shah2023gnm}. The GNM architecture consists of two convolutional encoders: one for the current observation image along with a history of $5$ previous images, and the other for the goal image (fused channel-wise with the other inputs). The output of these two encoders is concatenated and decoded using an MLP and two linear projection heads to predict: a) \textit{local waypoints}, that is, position and yaw relative to the current robot position, and b) \textit{distance} to the image goal. Our object-relative controller uses a modified version of this architecture with a \textit{single} goal encoder for the WayObject Costmap, which is based on a custom ResNet model with $D$ input channels. Our WayObject Costmap has the same resolution as the images used by GNM ($W=85$, $H=64$), but each pixel is represented by its locally-normalized path length encoding $E$ (see Equation~\ref{eq:encoding}). We predict a trajectory rollout of $10$ future 2D waypoints in the BEV space of the robot's frame of reference. As our object-relative encoders do not require an explicit goal image, we ignore the distance prediction output.

\subsection{Mapping} We used the automatic mask generator of SAM2~\cite{ravi2025sam2} (ViT-L) to obtain segmentation masks, sampling $16$ points per side and using $1$ additional cropping layer with a downsample factor of $2$. We used CLIP~\cite{radford2021learning} to remove segments if their masked image embedding matched with text string `floor' or `ceiling'. We used Superpoint-Lightglue~\cite{lindenberger2023lightglue} to match every current frame with $3$ previous frames to obtain keypoint correspondences. To enable batching, we used a fixed number of SuperPoint keypoints ($2048$) and disabled early stopping and point pruning in LightGlue. To obtain a local 3D projection of a segment, we used the indoor model checkpoint from Depth Anything~\cite{depthanything}. 

\subsection{Localization and Planning} 
\label{sec_app:loc_plan}
We used FastSAM~\cite{zhao2023fast} from Ultralytics for segmentation during the execution phase, using a mask confidence threshold of $0.5$. For localization, we used a submap centered at the reference map image that is closest to the agent's current state in the simulator. We used a radius of $16$ frames with a subsampling factor of $2$ to obtain the submap images against which the query image is localized through pairwise image matching (Superpoint-LightGlue~\cite{lindenberger2023lightglue}). For real-world experiments, the submap center is calculated as the reference map image that has the most segments matched with the current query image and its history of $8$ frames. For efficient global path planning, we precompute all path lengths from all the map nodes to the goal node. The goal node is estimated based on the maximum IoU of the predicted segments with the ground truth segmentation mask of the goal object -- if IoU=0, the episode is deemed a failure, though it occurs rarely. For each matched segment pair between the query and a reference map image, we look-up the path length from the precomputed data. As some of the query segments get matched with multiple map nodes, we only consider the match that has the least path length. For the query segments that could not be localized, we assign them a path length through a \textbf{tracking mechanism}: we perform pairwise image matching between the current image and its history of $8$ frames, and use these segment correspondences to track path lengths and assign a median path length to the current query segment that was not localized but could be tracked. Finally, any undetected, or unlocalized and untracked, image segments are marked as outliers and encoded with a $0$ value, as described in Section~\ref{sec_app:costmap}. 

\subsection{Control Execution} We broadly follow GNM~\cite{shah2023gnm} in terms of execution of the predicted waypoints. We used the last waypoint from the predicted trajectory rollout to calculate linear and angular velocities. We clip the linear velocity between $0$ and $0.05$ $m/s$, and the angular velocity between $-0.1$ and $0.1$ $rad/s$. We compute a moving window average of these velocity values over predictions from the past $5$ image observations. 
For the real-world experiments, we clip the linear velocity between $0.15$ to $0.4$ $m/s$; we execute the predicted velocities on the Unitree Go1 robot dog using its \texttt{HighCmd} control interface.

\subsection{Parameters: Image resolution, field-of-view and agent radius} For our simulator experiments, we consistently used $120^\circ$ field-of-view images. For the mapping and execution phases, we used an agent radius of $0.75m$ and an image resolution of $320\times240$. For controller training, we used an agent radius of $0.3m$, and an image resolution of $85\times65$ -- the same as that used in GNM~\cite{shah2023gnm}.

\section{Experimental Setup}
In this section, we provide details of the baseline methods. We also discuss the episodes which were excluded during the evaluation, which highlights the limitations and implementation overheads of the underlying simulator.

\subsection{Baselines}
\label{sec_app:baselines}
\textbf{PixNav~\cite{cai2024bridging}:} PixNav is a transformer-based imitation learning method for local navigation~\cite{cai2024bridging}, which utilizes a patch of goal pixels representing either the final destination or intermediate navigation targets. This goal patch is initially input to the model as a mask alongside the corresponding RGB image, and the model then selects an action from a discrete set: \{Stop, MoveAhead, TurnLeft, TurnRight, LookUp, LookDown\}. At each following step, the model uses the current RGB image, a collision signal, the sequence of previous images, and the initial goal mask to predict the next action. Unlike RoboHop and our proposed method ObjectReact, which are continuous controllers with fixed cameras, PixNav operates as a discrete controller with a movable camera. To accommodate these differences and align with PixNav's design, its evaluations were configured to begin with a visible intermediate goal determined by our global path planner. The goal is then updated either when the model outputs the `Done' command or when the memory buffer reached capacity. We used the official model checkpoint provided by the original authors.

\textbf{RoboHop~\cite{RoboHop}:} We use its zero-shot controller based on path length-weighted 2D pixel centers of the object segments:
\begin{equation}
\Delta \phi = \frac{g}{w} \sum_j \alpha_j(m_j - o)
\label{eq:robohop}; \quad \alpha_j = \frac{e^{\beta d_j}}{\sum_j e^{\beta d_j}}
\end{equation}
where $m_j$ denotes the centroid of segment $S_j$, and $o$ represents the image center. Each $d_j$ is the min-max normalized path length associated with segment $S_j$. The weights $\alpha_j$ are computed using a softmax function with temperature parameter $\beta=5$. The term $w$ refers to the image width, and $g$ is a fixed gain constant set to $0.4$.

\subsection{Excluded episodes in evaluation}
\label{sec_app:excluded_episodes}
We procedurally generate object goals for the challenging tasks of Alt-Goal, Reverse, and Shortcut. Through manual inspection, we then exclude episodes for a task in cases where the goal is invalid, for example, when the goal is unreachable due to incorrect rendering of the surrounding obstacles or when it covers a large part of the image (e.g., when only a wall is visible). We also exclude episodes where the goal does not meet the criteria of the task, for example, when all possible alternative goals for an episode fall on the path to the original goal. Finally, we exclude scenes with rendering issues such as \textit{i)} large areas of missing geometry and \textit{ii)} areas that are visually traversable but not reachable in the simulator. This leads to a removal of $6$, $16$, $13$, and $9$ episodes for the Imitate, Alt-Goal, Shortcut, and Reverse tasks respectively, out of $36$ original episodes per task.

\section{Additional Results}

\subsection{Quantitative Analyses}
\label{sec_app:quant}
\textbf{Comparing different GNM models:} In the main paper, we used the GNM model representing an image-relative approach, which we trained on the same simulator data as that used for our proposed object-relative method ObjectReact. Here, we include results for the GNM model that was trained on a variety of real-world outdoor scenes and embodiments, using the official checkpoint provided by the original authors. As observed in Table~\ref{supp_res:gnm}, both the GNM models -- sim and real world -- perform similarly; this shows that the performance gap on the challenging tasks is not attributed to the training environment or its diversity in terms of embodiment, but instead on the type of world representation (image- versus object-level) and the subgoal conditioning type (image pairs versus WayObject costmap). 

\textbf{Scalability of offline mapping and planning}
We perform planning offline as the map is assumed to be given. We observed that map/plan time scales roughly linearly with the number of images (or object nodes). In Table~\ref{tab:scalability} presents compute time of different modules for an increasing map size, showcasing that a three-floor, $30 \times 36$m office building can be mapped in $5$ minutes (including relatively cluttered desks etc.). If mapping were to be performed online, this cost would be amortized throughout the run.

\begin{table}
    \centering
    \caption{Scalability analyses of offline mapping and planning on a real-world office building. }
    \begin{adjustbox}{width=0.9\textwidth}
    \begin{tabular}{l l c c c}
    \toprule
         & \textbf{Method} & \textbf{One floor} & \textbf{Two floors} & \textbf{Three floors} \\

        \midrule

        Num. Images / Nodes & & 171 / 3521 & 417 / 7959 & 639 / 10252 \\
        \midrule

        {Segmentation} & FastSAM~\cite{zhao2023fast} & 18.5s & 41.5s & 63.2s \\
        {Matching} & SuperPoint-LightGlue~\cite{lindenberger2023lightglue} & 40.1s & 94.6s & 146.9s \\
        {Depth estimation} & Depth Anything~\cite{depthanything} & 14.5s & 34.0s & 51.7s \\
        {Planning} & Dijkstra's Algorithm~\cite{dijksta1959note, hagberg2008exploring} & 37ms & 96ms & 112ms \\

    \bottomrule
    \end{tabular}
    \end{adjustbox}
    \label{tab:scalability}
\end{table}

\subsection{Qualitative Results and Analyses}
\begin{figure*}
    \centering
    \includegraphics[width=0.75\textwidth]{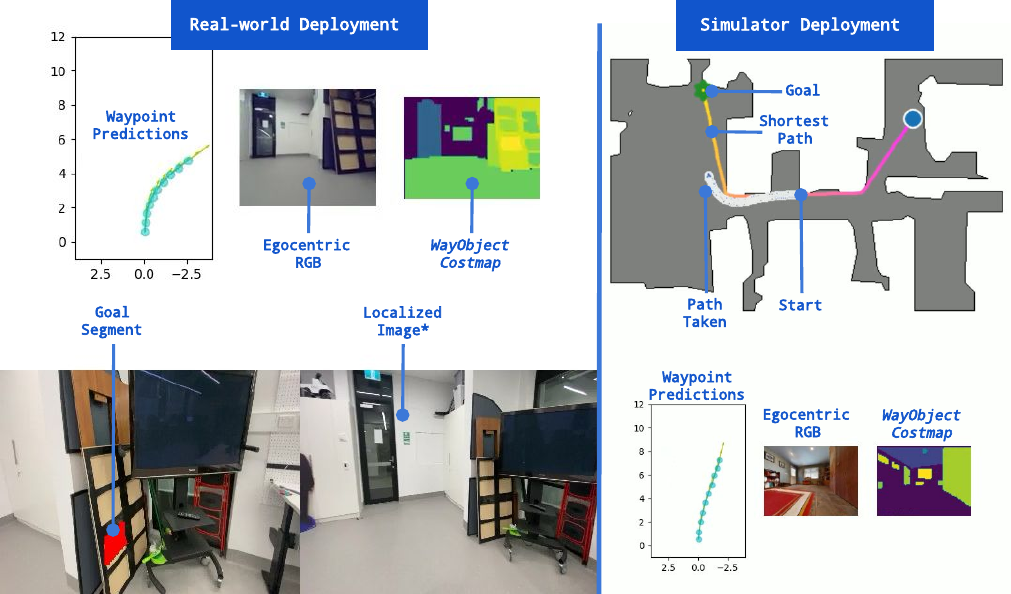}
    \caption{\textbf{Examples of demonstration videos}. Real-world demonstration video example (left) and simulator video example (right). \textbf{*} The localized image is the closest match found in the map (see Section \ref{sec_app:loc_plan} for details).}
    \label{fig:visualization_examples}
\end{figure*}

We provide details of our real-world and simulated experiments, with reference to the accompanying demonstration videos. Figure~\ref{fig:visualization_examples} provides details of the visualization panel of these videos. We tested navigation capabilities on the Unitree Go1 robot dog~\cite{unitree_go1} on a variety of tasks and environmental conditions. In the following, we discuss many successful trials that show that object-relative navigation is a promising avenue for further research. 

\begin{table}
\centering
\caption{Comparing (object-relative) ObjectReact against (image-relative) GNM models, trained on real world and simulator data.}
\begin{adjustbox}{width=\textwidth}

\begin{tabular}{lc cc cc cc cc} 
\toprule
    & &\multicolumn{2}{c}{\textbf{Imitate}} & \multicolumn{2}{c}{\textbf{Alt Goal}} & \multicolumn{2}{c}{\textbf{Shortcut}} & \multicolumn{2}{c}{\textbf{Reverse}} \\

\cmidrule(lr{0.75em}){3-10}

  Method & Train Data   & SPL & SSPL & SPL & SSPL &SPL & SSPL & SPL & SSPL \\

\cmidrule(lr{0.75em}){1-1}
\cmidrule(lr{0.75em}){2-2}
\cmidrule(lr{0.75em}){3-4}
\cmidrule(lr{0.75em}){5-6}
\cmidrule(lr{0.75em}){7-8}
\cmidrule(lr{0.75em}){9-10}

 GNM & Sim & 57.58 & \textbf{66.15} & 2.17 &  13.54 & 7.69 &  18.40 & 11.60 & 23.59 \\
 GNM & Real & 48.48 & 56.18 & 6.52 &  17.08 & 9.62 &  24.05 & 3.33 &  8.54 \\

 ObjectReact (Ours) & Sim & \textbf{59.08} & 64.62 & \textbf{21.74} & \textbf{27.40 }& \textbf{23.08} & \textbf{39.56} & \textbf{26.67} & \textbf{36.69} \\

\bottomrule
\end{tabular}
\label{supp_res:gnm}
\end{adjustbox}
\end{table}

\subsubsection{Simulator Deployment}

We have included several example videos of successful runs for our ObjectReact agent deployed in the simulator. These runs do not use any ground truth object instance information: FastSAM is used for segmentation and SuperPoint-LightGlue for matching. We have additionally included a deployment run that uses ground-truth instances, \texttt{ground\_truth\_perception\_imitate\_sim.mp4}, to highlight the gap in perception quality that ObjectReact must overcome during deployment using the inference-based perception pipeline. We also include several failed episodes that demonstrate failure to avoid objects (e.g. \texttt{imitate\_failure\_sim.mp4}), and perception failures when facing away from the goal or in a position that diverges from the mapping run (e.g. 
\texttt{reverse\_failure\_sim.mp4}). Another failure mode is inability to match the goal segment in the map. This is more common for Alt Goal and Reverse tasks, where the goal may be on the periphery of the mapping perspective.

\subsubsection{Real-World Deployment}

Here we provide a qualitative overview of the deployment of ObjectReact on the Unitree Go1 quadruped robot under various conditions. For each trial, we include the video we used to generate the map as well as the robot's egocentric observations during autonomous deployment. 

\begin{figure}
    \centering
    \includegraphics[width=0.8\textwidth]{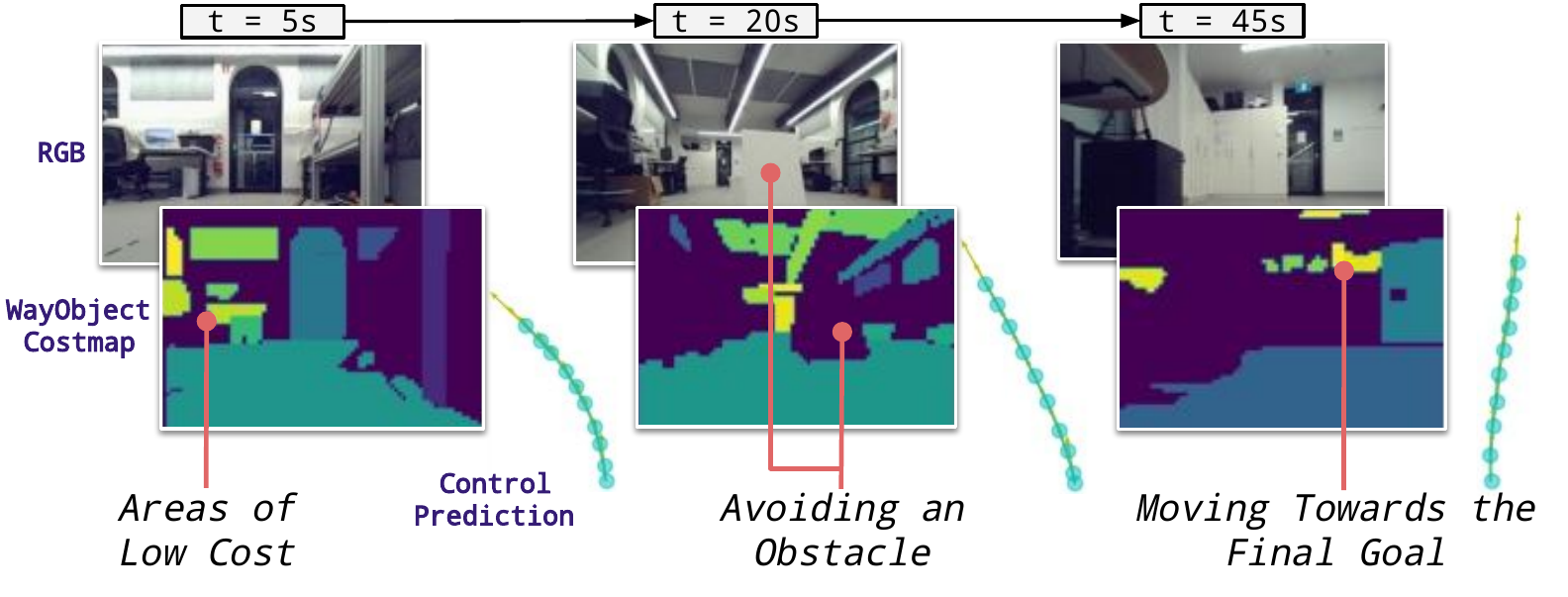}
    \caption{\textbf{Real-world Experiments}. We deploy our approach on the Unitree Go1 robot dog \cite{unitree_go1}. Here, we show egocentric RGB images, their corresponding WayObject Costmaps, and the predicted trajectory rollout at several timesteps during autonomous navigation to the goal object.
    At $t = 5s$, the policy chooses to turn left towards a region of lower-cost objects. At $t = 20s$, it successfully navigates around an obstacle (here visible as a region of low-cost). Finally, at $t = 45s$, it moves towards the final goal, subsequently succeeding at reaching the goal object.
    }
    \label{fig:real-world-experiments}
\end{figure}

\textbf{New Obstacle in the Map:} 
Figure~\ref{fig:real-world-experiments} illustrates that despite sim-only training, the robot is able to successfully navigate using our WayObject Costmap-conditioned learnt controller. At $t = 5s$, the robot makes a left turn towards an area of lower cost. Part-way through, it navigates around an obstacle (a region of high cost in the Costmap), and finally reaches the goal object. The map trajectory for this experiment did not have any obstacles, thus the results here highlight a re-routing capability when the map has changed. This ability mainly stems from accurate data association which leaves the unseen node as unmatched, thus leading to a high cost as an outlier.

\textbf{Cross-embodiment generalisation:} We examined cross-embodiment generalization between mapping and execution in the real world. We used maps generated from videos taken using a \textit{\textbf{phone camera}}, and deployed ObjectReact using these maps on the quadruped robot. We have included videos of a number of successful cross-embodiment demonstrations (those labeled \texttt{CrossEmbodiment} and several others). This generalization capability could be useful in future for taking advantage of multiple different sources of mapping data when a robot is navigating in a new or changing environment.

\textbf{Alternative tasks:} Here we examine the Alt Goal and Shortcut tasks. In \texttt{AltGoal\_humanoidRobot}, we demonstrate successful navigation towards an object that was only visible in the periphery of the mapping run. In \texttt{Shortcut\_cutout}, we show that ObjectReact is able to reach the goal along a direct path, even when the mapping run followed a much longer, winding path to the goal.

\textbf{Different environmental conditions:} We also investigated whether ObjectReact is robust to environmental changes between mapping and execution -- an expected advantage attributed to the use of open-set, zero-shot perception models. For \texttt{LowLight} trials, we generated the map with full lighting but deployed the robot under low lighting conditions. For \texttt{DayNight}, we generated the map under natural daylight conditions, but deployed during the evening. In both cases, ObjectReact was able to successfully generalize due to its WayObject Costmap representation that is largely invariant to these changes. In trials marked \texttt{Obstacles}, we also show that ObjectReact can also adapt to obstacles that were not present during the mapping run.

\section{Limitations}

\subsection{Perception} We presented results using two types of perception engines: one based on the ground truth instances from the simulator and the other based on SAM/FastSAM as the segmentor, and SuperPoint-LightGlue as the matcher. The simulator ground truth based results present an optimistic upper bound for our learnt controller under a perfect perception setting. In practice, when using inferred methods, although object-relative controller outperforms its image-relative counterpart, the absolute performance on the challenging tasks is still subpar. Through qualitative analyses, we found the following issues with the current state-of-the-art segmentors and matchers: a) SAM~\cite{kirillov2023segment} is computationally expensive, so we use it only for \textit{offline} mapping due to its high precision in producing consistent segmentations, however, we observed that it has low recall in terms of segmenting the entire image and often missed segmenting certain object instances, e.g., walls; b) FastSAM is highly efficient and we use it during the execution phase, however it often produces overlapping masks which also lack crisp boundaries; and c) SuperPoint-LightGlue~\cite{lindenberger2023lightglue} provide sparse local feature matches, which we observed to have better accuracy-speed trade-off in comparison to dense matchers such as RoMA~\cite{edstedt2024roma}; as noted in RoboHop~\cite{RoboHop}, the geometric verification of local feature matching is preferred over DINOv2-based segment matching for robustness to mismatches during localization. These findings highlight that accurate data association for image segments/objects still remains an open research challenge.

\subsection{Mapping and Planning} An interesting failure mode of our approach is its path planning through shortcuts created by instance categories like ceiling and floor. Due to the nature of their geometry, these object nodes usually have a high degree in the graph, which creates several shortcut paths and reduces the variance of the overall distribution of path lengths. Although we normalize path lengths per image when generating costmaps, the aforementioned issue is particularly intensified by a large margin between the low cost of localized segments and high outlier cost of unmatched or undetected segments. We address this issue by removing nodes from the map if its CLIP~\cite{radford2021learning} vision embedding matches with the text ``floor'' or ```ceiling''. Other possible solutions for removing such nodes could include depth- and robot' height-based plane estimation or using a cut-off on the degree of nodes in the graph. While this is only a symptomatic solution, the root issue of topologically planning across geometrically-variable object instances could be addressed by representing objects with an additional set of points that can capture its spatial extent or by using local 3D boundaries between neighboring objects when computing path lengths.

\subsection{Learnt Controller} We trained a local controller conditioned on high-level representation of subgoals in the form of WayObject Costmap. Unlike its image-relative counterpart, e.g., GNM~\cite{shah2023gnm}, we did not use any history context. This poses limitations when the costmap varies significantly from frame-to-frame due to potential failures in segmentation and matching. We observed that smoothing the costmap during the execution phase, by matching objects from the current observation with its history, stabilized the costmap. An alternative solution could be to learn these dynamics either through a stacked goal encoder similar to the history stacking of GNM, or by smoothing the goal encoding instead of the costmaps.

\begin{figure}
    \centering \includegraphics[width=\textwidth]{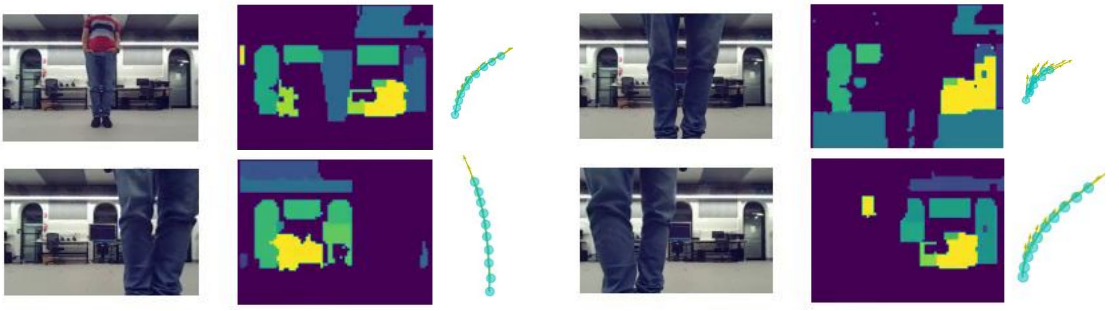}
    \caption{An illustration of the effect of dynamic objects and occlusions on the predicted control signal.}
    \label{fig:qual_dynamic}
\end{figure}

\subsection{Changes in the Map and Dynamic Objects}
Although minor pose changes of the existing objects (e.g., chairs) did not affect our real-world experiments, any significant scene rearrangement will likely cause failures due to misleading object-level connectivity of the precomputed map. Figure~\ref{fig:qual_dynamic} shows how the predicted trajectory varies with the dynamic object's motion, where failures mainly arise under severe occlusion and mismatching of dynamic obstacles. Note that the controller is \textbf{not} trained to avoid dynamic objects, \textit{this behavior simply emerges from the image matcher's true-negative detection ability}.

\subsection{Exploration}
In our experiments, we do not modify the initial map based on the observations the robot obtains during an episode. Ultimately, deployed approaches could use a controller that can explore the environment when the agent is lost or a goal cannot be identified in the map, while continuing to add to the map as new observations become available~\cite{shah2023vint, shah2022viking, sridhar2024nomad}. This would allow image-relative \textit{or} object-relative controllers to succeed at the proposed challenging tasks, provided they can explore efficiently. Our results indicate that WayObject Costmap representation is a suitable starting point for exploration, as the agent can still progress in a new region even when there are a very few objects matched against the map. 

\subsection{Real-world Deployment}
In our video demonstrations, we present a number of trials that show that our ObjectReact controller is able to generalize to real-world settings. However, there are several limitations of our approach that make real-world deployment challenging. Here, we outline these, enumerate common failure modes, and discuss our plans for overcoming these limitations in future.

\textbf{Training-deployment gap:} While our WayObject Costmap observations mitigate many problems of \textit{visual} generalization to the real world, there are several other challenges our approach must overcome when deployed on a real robot. First, camera parameters differ between the simulator training and real-world deployment. Second, the use of FastSAM + LightGlue in deployment for perception leads to different kinds of perception failures in comparison to those observed during training (which uses ground-truth simulator perception with data augmentation based on perturbation of path lengths). Finally, as shortest path demonstrations are used during training, cloning this behaviour results in an agent that prefers to closely corner around objects, leaving little room for error; also, the demonstrations go directly to the goal, lacking examples of taking slightly longer routes around obstacles.

\textbf{Failure modes:} The above issues result in several common failure modes during real-world operation. These included \textbf{a)} collisions, \textbf{b)} perception failures, and \textbf{c)} deviating from the mapping area. Because the ObjectReact policy is trained to take the shortest path to the goal, it often leaves little margin for error when navigating past obstacles, sometimes resulting in collisions during execution (see \texttt{failure\_close\_to\_obstacle.mp4}). Perception failures can occur when the perception pipeline fails to match important segments (like the goal or an obstacle), or matches a segment that connects to many other objects, resulting in WayObject Costmaps that have large, low-cost regions that are difficult to differentiate. During deployment, the agent can also deviate from the mapping area due to perception failures etc., but recovery from this state is not explicitly demonstrated in the training data (see \texttt{failure\_diverged\_from\_map\_matched\_wall.mp4}).

We hope to address the above limitations in future by \textbf{a)} incorporating real-world training data, \textbf{b)} using a demonstration path-finding algorithm that leaves a larger margin for obstacles where possible, \textbf{c)} training with the inferred perception pipeline to minimize sim-to-real gap, and \textbf{d)} further improving the perception pipeline.

\clearpage
\bibliography{references}

\begin{thebibliography}{93}
\providecommand{\natexlab}[1]{#1}
\providecommand{\url}[1]{\texttt{#1}}
\expandafter\ifx\csname urlstyle\endcsname\relax
  \providecommand{\doi}[1]{doi: #1}\else
  \providecommand{\doi}{doi: \begingroup \urlstyle{rm}\Url}\fi

\bibitem[Savinov et~al.(2018)Savinov, Dosovitskiy, and Koltun]{savinov2018semi}
N.~Savinov, A.~Dosovitskiy, and V.~Koltun.
\newblock Semi-parametric topological memory for navigation.
\newblock In \emph{International Conference on Learning Representations}, 2018.

\bibitem[Shah et~al.(2023{\natexlab{a}})Shah, Sridhar, Dashora, Stachowicz, Black, Hirose, and Levine]{shah2023vint}
D.~Shah, A.~Sridhar, N.~Dashora, K.~Stachowicz, K.~Black, N.~Hirose, and S.~Levine.
\newblock Vint: A large-scale, multi-task visual navigation backbone with cross-robot generalization.
\newblock In \emph{7th Annual Conference on Robot Learning}, 2023{\natexlab{a}}.

\bibitem[Shah et~al.(2023{\natexlab{b}})Shah, Sridhar, Bhorkar, Hirose, and Levine]{shah2023gnm}
D.~Shah, A.~Sridhar, A.~Bhorkar, N.~Hirose, and S.~Levine.
\newblock Gnm: A general navigation model to drive any robot.
\newblock In \emph{2023 IEEE International Conference on Robotics and Automation (ICRA)}, pages 7226--7233. IEEE, 2023{\natexlab{b}}.

\bibitem[Chaplot et~al.(2020)Chaplot, Salakhutdinov, Gupta, and Gupta]{chaplot2020neural}
D.~S. Chaplot, R.~Salakhutdinov, A.~Gupta, and S.~Gupta.
\newblock Neural topological slam for visual navigation.
\newblock In \emph{Proceedings of the IEEE/CVF Conference on Computer Vision and Pattern Recognition}, pages 12875--12884, 2020.

\bibitem[Gu et~al.(2023)Gu, Kuwajerwala, Morin, Jatavallabhula, Sen, Agarwal, Rivera, Paul, Ellis, Chellappa, Gan, {de Melo}, Tenenbaum, Torralba, Shkurti, and Paull]{conceptgraphs}
Q.~Gu, A.~Kuwajerwala, S.~Morin, K.~Jatavallabhula, B.~Sen, A.~Agarwal, C.~Rivera, W.~Paul, K.~Ellis, R.~Chellappa, C.~Gan, C.~{de Melo}, J.~Tenenbaum, A.~Torralba, F.~Shkurti, and L.~Paull.
\newblock Conceptgraphs: Open-vocabulary 3d scene graphs for perception and planning.
\newblock In \emph{arXiv}, 2023.

\bibitem[Salas-Moreno et~al.(2013)Salas-Moreno, Newcombe, Strasdat, Kelly, and Davison]{salas2013slam++}
R.~F. Salas-Moreno, R.~A. Newcombe, H.~Strasdat, P.~H. Kelly, and A.~J. Davison.
\newblock Slam++: Simultaneous localisation and mapping at the level of objects.
\newblock In \emph{Proceedings of the IEEE conference on computer vision and pattern recognition}, pages 1352--1359, 2013.

\bibitem[Georgakis et~al.(2022)Georgakis, Bucher, Schmeckpeper, Singh, and Daniilidis]{georgakis2022learning}
G.~Georgakis, B.~Bucher, K.~Schmeckpeper, S.~Singh, and K.~Daniilidis.
\newblock Learning to map for active semantic goal navigation.
\newblock In \emph{The Tenth International Conference on Learning Representations (ICLR 2022)}, 2022.

\bibitem[Paul et~al.(2024)Paul, Garg, Choudhary, Singh, and Krishna]{paul2024lego}
P.~Paul, A.~Garg, T.~Choudhary, A.~K. Singh, and K.~M. Krishna.
\newblock Lego-drive: Language-enhanced goal-oriented closed-loop end-to-end autonomous driving.
\newblock In \emph{2024 IEEE/RSJ International Conference on Intelligent Robots and Systems (IROS)}, pages 10020--10026. IEEE, 2024.

\bibitem[Huang et~al.(2023)Huang, Mees, Zeng, and Burgard]{huang23vlmaps}
C.~Huang, O.~Mees, A.~Zeng, and W.~Burgard.
\newblock Visual language maps for robot navigation.
\newblock In \emph{Proceedings of the IEEE International Conference on Robotics and Automation (ICRA)}, London, UK, 2023.

\bibitem[Weiss et~al.(2011)Weiss, Scaramuzza, and Siegwart]{weiss2011monocular}
S.~Weiss, D.~Scaramuzza, and R.~Siegwart.
\newblock Monocular-slam--based navigation for autonomous micro helicopters in gps-denied environments.
\newblock \emph{Journal of Field Robotics}, 28\penalty0 (6):\penalty0 854--874, 2011.

\bibitem[Zhang et~al.(2023)Zhang, Dai, Meng, Fan, Chen, Xu, and Wang]{zhang20233d}
J.~Zhang, L.~Dai, F.~Meng, Q.~Fan, X.~Chen, K.~Xu, and H.~Wang.
\newblock 3d-aware object goal navigation via simultaneous exploration and identification.
\newblock In \emph{Proceedings of the IEEE/CVF Conference on Computer Vision and Pattern Recognition}, pages 6672--6682, 2023.

\bibitem[Zhao et~al.(2023)Zhao, Zhang, He, Qiao, and Liu]{zhao2023zero}
Q.~Zhao, L.~Zhang, B.~He, H.~Qiao, and Z.~Liu.
\newblock Zero-shot object goal visual navigation.
\newblock In \emph{2023 IEEE International Conference on Robotics and Automation (ICRA)}, pages 2025--2031. IEEE, 2023.

\bibitem[Dorbala et~al.(2023)Dorbala, Mullen~Jr, and Manocha]{dorbala2023can}
V.~S. Dorbala, J.~F. Mullen~Jr, and D.~Manocha.
\newblock Can an embodied agent find your “cat-shaped mug”? llm-based zero-shot object navigation.
\newblock \emph{IEEE Robotics and Automation Letters}, 2023.

\bibitem[Chaplot et~al.(2020)Chaplot, Gandhi, Gupta, and Salakhutdinov]{chaplot2020object}
D.~S. Chaplot, D.~P. Gandhi, A.~Gupta, and R.~R. Salakhutdinov.
\newblock Object goal navigation using goal-oriented semantic exploration.
\newblock \emph{Advances in Neural Information Processing Systems}, 33:\penalty0 4247--4258, 2020.

\bibitem[Kim et~al.(2022)Kim, Kwon, Yoo, Choi, Park, and Oh]{TSGM}
N.~Kim, O.~Kwon, H.~Yoo, Y.~Choi, J.~Park, and S.~Oh.
\newblock {Topological Semantic Graph Memory for Image Goal Navigation}.
\newblock In \emph{CoRL}, 2022.

\bibitem[Kwon et~al.(2021)Kwon, Kim, Choi, Yoo, Park, and Oh]{kwon2021visual}
O.~Kwon, N.~Kim, Y.~Choi, H.~Yoo, J.~Park, and S.~Oh.
\newblock Visual graph memory with unsupervised representation for visual navigation.
\newblock In \emph{Proceedings of the IEEE/CVF international conference on computer vision}, pages 15890--15899, 2021.

\bibitem[Armeni et~al.(2019)Armeni, He, Gwak, Zamir, Fischer, Malik, and Savarese]{armeni20193d}
I.~Armeni, Z.-Y. He, J.~Gwak, A.~R. Zamir, M.~Fischer, J.~Malik, and S.~Savarese.
\newblock 3d scene graph: A structure for unified semantics, 3d space, and camera.
\newblock In \emph{Proceedings of the IEEE/CVF international conference on computer vision}, pages 5664--5673, 2019.

\bibitem[Rosinol et~al.(2021)Rosinol, Violette, Abate, Hughes, Chang, Shi, Gupta, and Carlone]{rosinol2021kimera}
A.~Rosinol, A.~Violette, M.~Abate, N.~Hughes, Y.~Chang, J.~Shi, A.~Gupta, and L.~Carlone.
\newblock Kimera: From slam to spatial perception with 3d dynamic scene graphs.
\newblock \emph{The International Journal of Robotics Research}, 40\penalty0 (12-14):\penalty0 1510--1546, 2021.

\bibitem[Rana et~al.(2023)Rana, Haviland, Garg, Abou-Chakra, Reid, and Suenderhauf]{rana2023sayplan}
K.~Rana, J.~Haviland, S.~Garg, J.~Abou-Chakra, I.~Reid, and N.~Suenderhauf.
\newblock Sayplan: Grounding large language models using 3d scene graphs for scalable task planning.
\newblock In \emph{7th Annual Conference on Robot Learning}, 2023.
\newblock URL \url{https://openreview.net/forum?id=wMpOMO0Ss7a}.

\bibitem[Werby et~al.(2024)Werby, Huang, B{\"u}chner, Valada, and Burgard]{werby2024hierarchical}
A.~Werby, C.~Huang, M.~B{\"u}chner, A.~Valada, and W.~Burgard.
\newblock Hierarchical open-vocabulary 3d scene graphs for language-grounded robot navigation.
\newblock In \emph{First Workshop on Vision-Language Models for Navigation and Manipulation at ICRA 2024}, 2024.

\bibitem[Ravichandran et~al.(2022)Ravichandran, Peng, Hughes, Griffith, and Carlone]{ravichandran2022hierarchical}
Z.~Ravichandran, L.~Peng, N.~Hughes, J.~D. Griffith, and L.~Carlone.
\newblock Hierarchical representations and explicit memory: Learning effective navigation policies on 3d scene graphs using graph neural networks.
\newblock In \emph{2022 International Conference on Robotics and Automation (ICRA)}, pages 9272--9279. IEEE, 2022.

\bibitem[Seymour et~al.(2022)Seymour, Mithun, Chiu, Samarasekera, and Kumar]{seymour2022graphmapper}
Z.~Seymour, N.~C. Mithun, H.-P. Chiu, S.~Samarasekera, and R.~Kumar.
\newblock Graphmapper: Efficient visual navigation by scene graph generation.
\newblock In \emph{2022 26th International Conference on Pattern Recognition (ICPR)}, pages 4146--4153. IEEE, 2022.

\bibitem[Singh et~al.(2023)Singh, Salvador, Weihs, and Kembhavi]{singh2023scene}
K.~P. Singh, J.~Salvador, L.~Weihs, and A.~Kembhavi.
\newblock Scene graph contrastive learning for embodied navigation.
\newblock In \emph{Proceedings of the IEEE/CVF International Conference on Computer Vision}, pages 10884--10894, 2023.

\bibitem[Yin et~al.(2024)Yin, Xu, Wu, Zhou, and Lu]{yin2024sg}
H.~Yin, X.~Xu, Z.~Wu, J.~Zhou, and J.~Lu.
\newblock Sg-nav: Online 3d scene graph prompting for llm-based zero-shot object navigation.
\newblock \emph{Advances in neural information processing systems}, 37:\penalty0 5285--5307, 2024.

\bibitem[Liu et~al.(2023)Liu, Wang, Wang, and Yang]{liu2023bird}
R.~Liu, X.~Wang, W.~Wang, and Y.~Yang.
\newblock Bird's-eye-view scene graph for vision-language navigation.
\newblock In \emph{Proceedings of the IEEE/CVF International Conference on Computer Vision}, pages 10968--10980, 2023.

\bibitem[Shah and Levine(2022)]{shah2022viking}
D.~Shah and S.~Levine.
\newblock {ViKiNG: Vision-Based Kilometer-Scale Navigation with Geographic Hints}.
\newblock In \emph{Proceedings of Robotics: Science and Systems}, 2022.
\newblock URL \url{http://www.roboticsproceedings.org/rss18/p019.html}.

\bibitem[Shah et~al.(2022)Shah, Osinski, Ichter, and Levine]{shah2022lmnav}
D.~Shah, B.~Osinski, B.~Ichter, and S.~Levine.
\newblock {LM}-nav: Robotic navigation with large pre-trained models of language, vision, and action.
\newblock In \emph{6th Annual Conference on Robot Learning}, 2022.
\newblock URL \url{https://openreview.net/forum?id=UW5A3SweAH}.

\bibitem[Sridhar et~al.(2024)Sridhar, Shah, Glossop, and Levine]{sridhar2024nomad}
A.~Sridhar, D.~Shah, C.~Glossop, and S.~Levine.
\newblock Nomad: Goal masked diffusion policies for navigation and exploration.
\newblock In \emph{2024 IEEE International Conference on Robotics and Automation (ICRA)}, pages 63--70. IEEE, 2024.

\bibitem[Shah et~al.(2021)Shah, Eysenbach, Kahn, Rhinehart, and Levine]{shah2021ving}
D.~Shah, B.~Eysenbach, G.~Kahn, N.~Rhinehart, and S.~Levine.
\newblock Ving: Learning open-world navigation with visual goals.
\newblock In \emph{2021 IEEE International Conference on Robotics and Automation (ICRA)}, pages 13215--13222. IEEE, 2021.

\bibitem[Shah et~al.(2022)Shah, Eysenbach, Rhinehart, and Levine]{shah2022rapid}
D.~Shah, B.~Eysenbach, N.~Rhinehart, and S.~Levine.
\newblock Rapid exploration for open-world navigation with latent goal models.
\newblock In \emph{Conference on Robot Learning}, pages 674--684. PMLR, 2022.

\bibitem[Sutton(1988)]{sutton1988learning}
R.~S. Sutton.
\newblock Learning to predict by the methods of temporal differences.
\newblock \emph{Machine learning}, 3\penalty0 (1):\penalty0 9--44, 1988.

\bibitem[Horswill(1993)]{horswill1993polly}
I.~Horswill.
\newblock Polly: A vision-based artificial agent.
\newblock In \emph{AAAI}, pages 824--829, 1993.

\bibitem[Matsumoto et~al.(1996)Matsumoto, Inaba, and Inoue]{matsumoto1996visual}
Y.~Matsumoto, M.~Inaba, and H.~Inoue.
\newblock Visual navigation using view-sequenced route representation.
\newblock In \emph{Proceedings of IEEE International conference on Robotics and Automation}, volume~1, pages 83--88. IEEE, 1996.

\bibitem[Vassallo et~al.(2000)Vassallo, Schneebeli, and Santos-Victor]{vassallo2000visual}
R.~F. Vassallo, H.~J. Schneebeli, and J.~Santos-Victor.
\newblock Visual servoing and appearance for navigation.
\newblock \emph{Robotics and autonomous systems}, 31\penalty0 (1-2):\penalty0 87--97, 2000.

\bibitem[Thrun(1995)]{thrun1995approach}
S.~Thrun.
\newblock An approach to learning mobile robot navigation.
\newblock \emph{Robotics and Autonomous systems}, 15\penalty0 (4):\penalty0 301--319, 1995.

\bibitem[Saxena et~al.(2017)Saxena, Pandya, Kumar, Gaud, and Krishna]{saxena2017exploring}
A.~Saxena, H.~Pandya, G.~Kumar, A.~Gaud, and K.~M. Krishna.
\newblock Exploring convolutional networks for end-to-end visual servoing.
\newblock In \emph{2017 IEEE International Conference on Robotics and Automation (ICRA)}, pages 3817--3823. IEEE, 2017.

\bibitem[Pathak et~al.(2018)Pathak, Mahmoudieh, Luo, Agrawal, Chen, Shentu, Shelhamer, Malik, Efros, and Darrell]{pathak2018zero}
D.~Pathak, P.~Mahmoudieh, G.~Luo, P.~Agrawal, D.~Chen, Y.~Shentu, E.~Shelhamer, J.~Malik, A.~A. Efros, and T.~Darrell.
\newblock Zero-shot visual imitation.
\newblock In \emph{Proceedings of the IEEE conference on computer vision and pattern recognition workshops}, pages 2050--2053, 2018.

\bibitem[Li and Ko{\v{s}}ecka(2020)]{li2020learning}
Y.~Li and J.~Ko{\v{s}}ecka.
\newblock Learning view and target invariant visual servoing for navigation.
\newblock In \emph{2020 IEEE International Conference on Robotics and Automation (ICRA)}, pages 658--664. IEEE, 2020.

\bibitem[Meng et~al.(2020)Meng, Ratliff, Xiang, and Fox]{meng2020scaling}
X.~Meng, N.~Ratliff, Y.~Xiang, and D.~Fox.
\newblock Scaling local control to large-scale topological navigation.
\newblock In \emph{2020 IEEE International Conference on Robotics and Automation (ICRA)}, pages 672--678. IEEE, 2020.

\bibitem[Katara et~al.(2021)Katara, Harish, Pandya, Gupta, Sanchawala, Kumar, Bhowmick, and Krishna]{katara2021deepmpcvs}
P.~Katara, Y.~Harish, H.~Pandya, A.~Gupta, A.~Sanchawala, G.~Kumar, B.~Bhowmick, and M.~Krishna.
\newblock Deepmpcvs: Deep model predictive control for visual servoing.
\newblock In \emph{Conference on Robot Learning}, pages 2006--2015. PMLR, 2021.

\bibitem[Pathre et~al.(2024)Pathre, Gupta, Qureshi, Brunda, Brahmbhatt, and Krishna]{pathre2024imagine2servo}
P.~Pathre, G.~Gupta, M.~N. Qureshi, M.~Brunda, S.~Brahmbhatt, and K.~M. Krishna.
\newblock Imagine2servo: Intelligent visual servoing with diffusion-driven goal generation for robotic tasks.
\newblock In \emph{2024 IEEE/RSJ International Conference on Intelligent Robots and Systems (IROS)}, pages 13466--13472. IEEE, 2024.

\bibitem[Ehsani et~al.(2024)Ehsani, Gupta, Hendrix, Salvador, Weihs, Zeng, Singh, Kim, Han, Herrasti, et~al.]{ehsani2024spoc}
K.~Ehsani, T.~Gupta, R.~Hendrix, J.~Salvador, L.~Weihs, K.-H. Zeng, K.~P. Singh, Y.~Kim, W.~Han, A.~Herrasti, et~al.
\newblock Spoc: Imitating shortest paths in simulation enables effective navigation and manipulation in the real world.
\newblock In \emph{Proceedings of the IEEE/CVF Conference on Computer Vision and Pattern Recognition}, pages 16238--16250, 2024.

\bibitem[Hutchinson et~al.(1996)Hutchinson, Hager, and Corke]{hutchinson1996tutorial}
S.~Hutchinson, G.~D. Hager, and P.~I. Corke.
\newblock A tutorial on visual servo control.
\newblock \emph{IEEE transactions on robotics and automation}, 12\penalty0 (5):\penalty0 651--670, 1996.

\bibitem[Jones et~al.(1997)Jones, Andresen, and Crowley]{jones1997appearance}
S.~D. Jones, C.~Andresen, and J.~L. Crowley.
\newblock Appearance based process for visual navigation.
\newblock In \emph{Proceedings of the 1997 IEEE/RSJ International Conference on Intelligent Robot and Systems. Innovative Robotics for Real-World Applications. IROS'97}, volume~2, pages 551--557. IEEE, 1997.

\bibitem[Mezouar and Chaumette(2002)]{mezouar2002path}
Y.~Mezouar and F.~Chaumette.
\newblock Path planning for robust image-based control.
\newblock \emph{IEEE transactions on robotics and automation}, 18\penalty0 (4):\penalty0 534--549, 2002.

\bibitem[Blanc et~al.(2005)Blanc, Mezouar, and Martinet]{blanc2005indoor}
G.~Blanc, Y.~Mezouar, and P.~Martinet.
\newblock Indoor navigation of a wheeled mobile robot along visual routes.
\newblock In \emph{Proceedings of the 2005 IEEE international conference on robotics and automation}, pages 3354--3359. IEEE, 2005.

\bibitem[Remazeilles et~al.(2006)Remazeilles, Chaumette, and Gros]{remazeilles20063d}
A.~Remazeilles, F.~Chaumette, and P.~Gros.
\newblock 3d navigation based on a visual memory.
\newblock In \emph{Proceedings 2006 IEEE International Conference on Robotics and Automation, 2006. ICRA 2006.}, pages 2719--2725. IEEE, 2006.

\bibitem[Cherubini et~al.(2011)Cherubini, Chaumette, and Oriolo]{cherubini2011visual}
A.~Cherubini, F.~Chaumette, and G.~Oriolo.
\newblock Visual servoing for path reaching with nonholonomic robots.
\newblock \emph{Robotica}, 29\penalty0 (7):\penalty0 1037--1048, 2011.

\bibitem[Diosi et~al.(2011)Diosi, Segvic, Remazeilles, and Chaumette]{diosi2011experimental}
A.~Diosi, S.~Segvic, A.~Remazeilles, and F.~Chaumette.
\newblock Experimental evaluation of autonomous driving based on visual memory and image-based visual servoing.
\newblock \emph{IEEE Transactions on Intelligent Transportation Systems}, 12\penalty0 (3):\penalty0 870--883, 2011.

\bibitem[Bista et~al.(2016)Bista, Giordano, and Chaumette]{bista2016appearance}
S.~R. Bista, P.~R. Giordano, and F.~Chaumette.
\newblock Appearance-based indoor navigation by ibvs using line segments.
\newblock \emph{IEEE robotics and automation letters}, 1\penalty0 (1):\penalty0 423--430, 2016.

\bibitem[Ahmadi et~al.(2020)Ahmadi, Nardi, Chebrolu, and Stachniss]{ahmadi2020visual}
A.~Ahmadi, L.~Nardi, N.~Chebrolu, and C.~Stachniss.
\newblock Visual servoing-based navigation for monitoring row-crop fields.
\newblock In \emph{2020 IEEE International Conference on Robotics and Automation (ICRA)}, pages 4920--4926. IEEE, 2020.

\bibitem[Feng et~al.(2021)Feng, Wu, Zhao, and Vela]{feng2021trajectory}
S.~Feng, Z.~Wu, Y.~Zhao, and P.~A. Vela.
\newblock Trajectory servoing: Image-based trajectory tracking using slam.
\newblock \emph{CoRR}, 2021.

\bibitem[Garg et~al.(2024)Garg, Rana, Hosseinzadeh, Mares, Suenderhauf, Dayoub, and Reid]{RoboHop}
S.~Garg, K.~Rana, M.~Hosseinzadeh, L.~Mares, N.~Suenderhauf, F.~Dayoub, and I.~Reid.
\newblock Robohop: Segment-based topological map representation for open-world visual navigation.
\newblock In \emph{2024 International Conference on Robotics and Automation (ICRA)}. IEEE, 2024.

\bibitem[Cai et~al.(2024)Cai, Huang, Cheng, Long, Gao, Sun, and Dong]{cai2024bridging}
W.~Cai, S.~Huang, G.~Cheng, Y.~Long, P.~Gao, C.~Sun, and H.~Dong.
\newblock Bridging zero-shot object navigation and foundation models through pixel-guided navigation skill.
\newblock In \emph{2024 IEEE International Conference on Robotics and Automation (ICRA)}, pages 5228--5234. IEEE, 2024.

\bibitem[Podgorski et~al.(2025)Podgorski, Garg, Hosseinzadeh, Mares, Dayoub, and Reid]{podgorski2025tango}
S.~Podgorski, S.~Garg, M.~Hosseinzadeh, L.~Mares, F.~Dayoub, and I.~Reid.
\newblock Tango: Traversablility-aware navigation with local metric control for topological goals.
\newblock In \emph{2025 IEEE International Conference on Robotics and Automation (ICRA)}. IEEE, 2025.

\bibitem[Wu et~al.(2019)Wu, Wu, Tamar, Russell, Gkioxari, and Tian]{wu2019bayesian}
Y.~Wu, Y.~Wu, A.~Tamar, S.~Russell, G.~Gkioxari, and Y.~Tian.
\newblock Bayesian relational memory for semantic visual navigation.
\newblock In \emph{Proceedings of the IEEE/CVF international conference on computer vision}, pages 2769--2779, 2019.

\bibitem[Yang et~al.(2019)Yang, Wang, Farhadi, Gupta, and Mottaghi]{yang2018visual}
W.~Yang, X.~Wang, A.~Farhadi, A.~Gupta, and R.~Mottaghi.
\newblock Visual semantic navigation using scene priors.
\newblock In \emph{International Conference on Learning Representations}, 2019.

\bibitem[Du et~al.(2020)Du, Yu, and Zheng]{du2020learning}
H.~Du, X.~Yu, and L.~Zheng.
\newblock Learning object relation graph and tentative policy for visual navigation.
\newblock In \emph{European Conference on Computer Vision}, pages 19--34. Springer, 2020.

\bibitem[Yoo et~al.(2024)Yoo, Choi, Park, and Oh]{yoo2024commonsense}
H.~Yoo, Y.~Choi, J.~Park, and S.~Oh.
\newblock Commonsense-aware object value graph for object goal navigation.
\newblock \emph{IEEE Robotics and Automation Letters}, 9\penalty0 (5):\penalty0 4423--4430, 2024.

\bibitem[Hahn et~al.(2021)Hahn, Chaplot, Tulsiani, Mukadam, Rehg, and Gupta]{hahn2021no}
M.~Hahn, D.~S. Chaplot, S.~Tulsiani, M.~Mukadam, J.~M. Rehg, and A.~Gupta.
\newblock No rl, no simulation: Learning to navigate without navigating.
\newblock \emph{Advances in Neural Information Processing Systems}, 34:\penalty0 26661--26673, 2021.

\bibitem[Dijksta(1959)]{dijksta1959note}
E.~W. Dijksta.
\newblock A note on two problems in connexion with graphs.
\newblock \emph{Numerische mathematik}, 1\penalty0 (1):\penalty0 269--271, 1959.

\bibitem[Furgale and Barfoot(2010)]{furgale2010visual}
P.~Furgale and T.~D. Barfoot.
\newblock Visual teach and repeat for long-range rover autonomy.
\newblock \emph{Journal of field robotics}, 27\penalty0 (5):\penalty0 534--560, 2010.

\bibitem[{\v{S}}egvi{\'c} et~al.(2009){\v{S}}egvi{\'c}, Remazeilles, Diosi, and Chaumette]{vsegvic2009mapping}
S.~{\v{S}}egvi{\'c}, A.~Remazeilles, A.~Diosi, and F.~Chaumette.
\newblock A mapping and localization framework for scalable appearance-based navigation.
\newblock \emph{Computer Vision and Image Understanding}, 113\penalty0 (2):\penalty0 172--187, 2009.

\bibitem[Zhang and Kleeman(2009)]{zhang2009robust}
A.~M. Zhang and L.~Kleeman.
\newblock Robust appearance based visual route following for navigation in large-scale outdoor environments.
\newblock \emph{The International Journal of Robotics Research}, 28\penalty0 (3):\penalty0 331--356, 2009.

\bibitem[Dall’Osto et~al.(2021)Dall’Osto, Fischer, and Milford]{dall2021fast}
D.~Dall’Osto, T.~Fischer, and M.~Milford.
\newblock Fast and robust bio-inspired teach and repeat navigation.
\newblock In \emph{2021 IEEE/RSJ International Conference on Intelligent Robots and Systems (IROS)}, pages 500--507. IEEE, 2021.

\bibitem[Mattamala et~al.(2022)Mattamala, Chebrolu, and Fallon]{mattamala2022efficient}
M.~Mattamala, N.~Chebrolu, and M.~Fallon.
\newblock An efficient locally reactive controller for safe navigation in visual teach and repeat missions.
\newblock \emph{IEEE Robotics and Automation Letters}, 7\penalty0 (2):\penalty0 2353--2360, 2022.

\bibitem[Krajn{\'\i}k et~al.(2018)Krajn{\'\i}k, Majer, Halodov{\'a}, and Vintr]{krajnik2018navigation}
T.~Krajn{\'\i}k, F.~Majer, L.~Halodov{\'a}, and T.~Vintr.
\newblock Navigation without localisation: reliable teach and repeat based on the convergence theorem.
\newblock In \emph{2018 IEEE/RSJ International Conference on Intelligent Robots and Systems (IROS)}, pages 1657--1664. IEEE, 2018.

\bibitem[Halodov{\'a} et~al.(2019)Halodov{\'a}, Dvo{\v{r}}r{\'a}kov{\'a}, Majer, Vintr, Mozos, Dayoub, and Krajn{\'\i}k]{halodova2019predictive}
L.~Halodov{\'a}, E.~Dvo{\v{r}}r{\'a}kov{\'a}, F.~Majer, T.~Vintr, O.~M. Mozos, F.~Dayoub, and T.~Krajn{\'\i}k.
\newblock Predictive and adaptive maps for long-term visual navigation in changing environments.
\newblock In \emph{2019 IEEE/RSJ International Conference on Intelligent Robots and Systems (IROS)}, pages 7033--7039. IEEE, 2019.

\bibitem[Do et~al.(2019)Do, Carrillo-Arce, and Roumeliotis]{do2019high}
T.~Do, L.~C. Carrillo-Arce, and S.~I. Roumeliotis.
\newblock High-speed autonomous quadrotor navigation through visual and inertial paths.
\newblock \emph{The International Journal of Robotics Research}, 38\penalty0 (4):\penalty0 486--504, 2019.

\bibitem[Krajn{\'\i}k et~al.(2017)Krajn{\'\i}k, Crist{\'o}foris, Kusumam, Neubert, and Duckett]{krajnik2017image}
T.~Krajn{\'\i}k, P.~Crist{\'o}foris, K.~Kusumam, P.~Neubert, and T.~Duckett.
\newblock Image features for visual teach-and-repeat navigation in changing environments.
\newblock \emph{Robotics and Autonomous Systems}, 88:\penalty0 127--141, 2017.

\bibitem[Kumar et~al.(2018)Kumar, Gupta, Fouhey, Levine, and Malik]{kumar2018visual}
A.~Kumar, S.~Gupta, D.~Fouhey, S.~Levine, and J.~Malik.
\newblock Visual memory for robust path following.
\newblock \emph{Advances in neural information processing systems}, 31, 2018.

\bibitem[Levine and Shah(2023)]{levine2023learning}
S.~Levine and D.~Shah.
\newblock Learning robotic navigation from experience: principles, methods and recent results.
\newblock \emph{Philosophical Transactions of the Royal Society B}, 378\penalty0 (1869):\penalty0 20210447, 2023.

\bibitem[Kirillov et~al.(2023)Kirillov, Mintun, Ravi, Mao, Rolland, Gustafson, Xiao, Whitehead, Berg, Lo, et~al.]{kirillov2023segment}
A.~Kirillov, E.~Mintun, N.~Ravi, H.~Mao, C.~Rolland, L.~Gustafson, T.~Xiao, S.~Whitehead, A.~C. Berg, W.-Y. Lo, et~al.
\newblock Segment anything.
\newblock In \emph{Proceedings of the IEEE/CVF international conference on computer vision}, pages 4015--4026, 2023.

\bibitem[Ravi et~al.(2025)Ravi, Gabeur, Hu, Hu, Ryali, Ma, Khedr, R{\"a}dle, Rolland, Gustafson, et~al.]{ravi2025sam2}
N.~Ravi, V.~Gabeur, Y.-T. Hu, R.~Hu, C.~Ryali, T.~Ma, H.~Khedr, R.~R{\"a}dle, C.~Rolland, L.~Gustafson, et~al.
\newblock Sam 2: Segment anything in images and videos.
\newblock In \emph{The Thirteenth International Conference on Learning Representations}, 2025.

\bibitem[Zhao et~al.(2023)Zhao, Ding, An, Du, Yu, Li, Tang, and Wang]{zhao2023fast}
X.~Zhao, W.~Ding, Y.~An, Y.~Du, T.~Yu, M.~Li, M.~Tang, and J.~Wang.
\newblock Fast segment anything, 2023.

\bibitem[Lindenberger et~al.(2023)Lindenberger, Sarlin, and Pollefeys]{lindenberger2023lightglue}
P.~Lindenberger, P.-E. Sarlin, and M.~Pollefeys.
\newblock {LightGlue: Local Feature Matching at Light Speed}.
\newblock In \emph{ICCV}, 2023.

\bibitem[Yang et~al.(2024)Yang, Kang, Huang, Xu, Feng, and Zhao]{depthanything}
L.~Yang, B.~Kang, Z.~Huang, X.~Xu, J.~Feng, and H.~Zhao.
\newblock Depth anything: Unleashing the power of large-scale unlabeled data.
\newblock In \emph{CVPR}, 2024.

\bibitem[DeTone et~al.(2018)DeTone, Malisiewicz, and Rabinovich]{detone2018superpoint}
D.~DeTone, T.~Malisiewicz, and A.~Rabinovich.
\newblock Superpoint: Self-supervised interest point detection and description.
\newblock In \emph{Proceedings of the IEEE conference on computer vision and pattern recognition workshops}, pages 224--236, 2018.

\bibitem[Vaswani et~al.(2017)Vaswani, Shazeer, Parmar, Uszkoreit, Jones, Gomez, Kaiser, and Polosukhin]{vaswani2017attention}
A.~Vaswani, N.~Shazeer, N.~Parmar, J.~Uszkoreit, L.~Jones, A.~N. Gomez, {\L}.~Kaiser, and I.~Polosukhin.
\newblock Attention is all you need.
\newblock \emph{Advances in neural information processing systems}, 30, 2017.

\bibitem[Ramakrishnan et~al.(2021)Ramakrishnan, Gokaslan, Wijmans, Maksymets, Clegg, Turner, Undersander, Galuba, Westbury, Chang, et~al.]{Ramakrishnan2021HabitatMatterport3D}
S.~K. Ramakrishnan, A.~Gokaslan, E.~Wijmans, O.~Maksymets, A.~Clegg, J.~M. Turner, E.~Undersander, W.~Galuba, A.~Westbury, A.~X. Chang, et~al.
\newblock Habitat-matterport 3d dataset (hm3d): 1000 large-scale 3d environments for embodied ai.
\newblock In \emph{Thirty-fifth Conference on Neural Information Processing Systems Datasets and Benchmarks Track (Round 2)}, 2021.

\bibitem[Krantz et~al.(2022)Krantz, Lee, Malik, Batra, and Chaplot]{krantz2022instance}
J.~Krantz, S.~Lee, J.~Malik, D.~Batra, and D.~S. Chaplot.
\newblock Instance-specific image goal navigation: Training embodied agents to find object instances.
\newblock \emph{arXiv preprint arXiv:2211.15876}, 2022.

\bibitem[Yadav et~al.(2023)Yadav, Krantz, Ramrakhya, Ramakrishnan, Yang, Wang, Turner, Gokaslan, Berges, Mootaghi, Maksymets, Chang, Savva, Clegg, Chaplot, and Batra]{habitatchallenge2023}
K.~Yadav, J.~Krantz, R.~Ramrakhya, S.~K. Ramakrishnan, J.~Yang, A.~Wang, J.~Turner, A.~Gokaslan, V.-P. Berges, R.~Mootaghi, O.~Maksymets, A.~X. Chang, M.~Savva, A.~Clegg, D.~S. Chaplot, and D.~Batra.
\newblock Habitat challenge 2023.
\newblock \url{https://aihabitat.org/challenge/2023/}, 2023.

\bibitem[Anderson et~al.(2018)Anderson, Chang, Chaplot, Dosovitskiy, Gupta, Koltun, Kosecka, Malik, Mottaghi, Savva, et~al.]{anderson2018evaluation}
P.~Anderson, A.~Chang, D.~S. Chaplot, A.~Dosovitskiy, S.~Gupta, V.~Koltun, J.~Kosecka, J.~Malik, R.~Mottaghi, M.~Savva, et~al.
\newblock On evaluation of embodied navigation agents.
\newblock \emph{arXiv preprint arXiv:1807.06757}, 2018.

\bibitem[Datta et~al.(2021)Datta, Maksymets, Hoffman, Lee, Batra, and Parikh]{datta2021integrating}
S.~Datta, O.~Maksymets, J.~Hoffman, S.~Lee, D.~Batra, and D.~Parikh.
\newblock Integrating egocentric localization for more realistic point-goal navigation agents.
\newblock In J.~Kober, F.~Ramos, and C.~Tomlin, editors, \emph{Proceedings of the 2020 Conference on Robot Learning}, volume 155 of \emph{Proceedings of Machine Learning Research}, pages 313--328. PMLR, 16--18 Nov 2021.
\newblock URL \url{https://proceedings.mlr.press/v155/datta21a.html}.

\bibitem[Kemp et~al.(2022)Kemp, Edsinger, Clever, and Matulevich]{kemp2022designstretchcompactlightweight}
C.~C. Kemp, A.~Edsinger, H.~M. Clever, and B.~Matulevich.
\newblock The design of stretch: A compact, lightweight mobile manipulator for indoor human environments, 2022.
\newblock URL \url{https://arxiv.org/abs/2109.10892}.

\bibitem[Unitree(2025)]{unitree_go1}
Unitree.
\newblock Unitree go1 product page.
\newblock \url{https://www.unitree.com/go1}, 2025.
\newblock Accessed: 2025-05-01.

\bibitem[Garg et~al.(2018)Garg, Suenderhauf, and Milford]{garg2018lost}
S.~Garg, N.~Suenderhauf, and M.~Milford.
\newblock Lost? appearance-invariant place recognition for opposite viewpoints using visual semantics.
\newblock \emph{Robotics: Science and Systems XIV}, pages 1--10, 2018.

\bibitem[Keetha et~al.(2023)Keetha, Mishra, Karhade, Jatavallabhula, Scherer, Krishna, and Garg]{keetha2023anyloc}
N.~Keetha, A.~Mishra, J.~Karhade, K.~M. Jatavallabhula, S.~Scherer, M.~Krishna, and S.~Garg.
\newblock Anyloc: Towards universal visual place recognition.
\newblock \emph{IEEE Robotics and Automation Letters}, 9\penalty0 (2):\penalty0 1286--1293, 2023.

\bibitem[Garg et~al.(2024)Garg, Puligilla, Kolathaya, Krishna, and Garg]{garg24revisitanything}
K.~Garg, S.~S. Puligilla, S.~Kolathaya, M.~Krishna, and S.~Garg.
\newblock Revisit anything: Visual place recognition via image segment retrieval.
\newblock In \emph{European Conference on Computer Vision (ECCV)}, September 2024.

\bibitem[Foo et~al.(2005)Foo, Warren, Duchon, and Tarr]{foo2005humans}
P.~Foo, W.~H. Warren, A.~Duchon, and M.~J. Tarr.
\newblock Do humans integrate routes into a cognitive map? map-versus landmark-based navigation of novel shortcuts.
\newblock \emph{Journal of Experimental Psychology: Learning, Memory, and Cognition}, 31\penalty0 (2):\penalty0 195, 2005.

\bibitem[Radford et~al.(2021)Radford, Kim, Hallacy, Ramesh, Goh, Agarwal, Sastry, Askell, Mishkin, Clark, et~al.]{radford2021learning}
A.~Radford, J.~W. Kim, C.~Hallacy, A.~Ramesh, G.~Goh, S.~Agarwal, G.~Sastry, A.~Askell, P.~Mishkin, J.~Clark, et~al.
\newblock Learning transferable visual models from natural language supervision.
\newblock In \emph{International conference on machine learning}, pages 8748--8763. PMLR, 2021.

\bibitem[Hagberg et~al.(2008)Hagberg, Schult, and Swart]{hagberg2008exploring}
A.~A. Hagberg, D.~A. Schult, and P.~J. Swart.
\newblock Exploring network structure, dynamics, and function using networkx.
\newblock In \emph{Proceedings of the 7th Python in Science Conference}, pages 11 -- 15, Pasadena, CA USA, 2008.
\newblock URL \url{https://networkx.org/documentation/stable/index.html}.

\bibitem[Edstedt et~al.(2024)Edstedt, Sun, B{\"o}kman, Wadenb{\"a}ck, and Felsberg]{edstedt2024roma}
J.~Edstedt, Q.~Sun, G.~B{\"o}kman, M.~Wadenb{\"a}ck, and M.~Felsberg.
\newblock Roma: Robust dense feature matching.
\newblock In \emph{Proceedings of the IEEE/CVF Conference on Computer Vision and Pattern Recognition}, pages 19790--19800, 2024.

\end{thebibliography}

\end{document}